\documentclass[a4paper,fleqn]{cas-dc}

\pdfoutput=1

\usepackage[numbers]{natbib}
\usepackage{amssymb}
\usepackage{amsmath,amsfonts}
\usepackage{algorithmic}
\usepackage{algorithm}
\usepackage{array}
\usepackage[caption=false,font=normalsize,labelfont=sf,textfont=sf]{subfig}
\usepackage{textcomp}
\usepackage{stfloats}
\usepackage{url}
\usepackage{verbatim}
\usepackage{graphicx}
\usepackage{multirow}
\usepackage{pifont}
\usepackage{color} 
\usepackage{xcolor}
\hypersetup{
  colorlinks = true,
  linkcolor  = blue,
  filecolor  = blue, 
  urlcolor   = blue,
  citecolor  = blue,
  }
\usepackage{cite}
\usepackage{cleveref}
\usepackage{makecell}
\usepackage[font=small,labelfont=bf,labelsep=none]{caption}  
\usepackage{caption}
\usepackage{tabularx}
\def\tsc#1{\csdef{#1}{\textsc{\lowercase{#1}}\xspace}}
\tsc{WGM}
\tsc{QE}

\begin{document}

\let\WriteBookmarks\relax
\def\floatpagepagefraction{1}
\def\textpagefraction{.001}
\let\printorcid\relax 

\title[mode = title]{Hierarchical Matching and Reasoning for Multi-Query Image Retrieval}  

\author[1]{Zhong Ji} 
\ead{jizhong@tju.edu.cn}
\credit{Conceptualization, Writing - Original Draft, Writing - Review \& Editing, Resources, Funding acquisition, Supervision}
\author[1]{Zhihao Li} 
\ead{zh_li@tju.edu.cn}
\credit{Methodology, Software, Formal analysis, Writing - Original Draft, Writing - Review \& Editing}
\author[1]{Yan Zhang} 
\cormark[1]
\ead{yzhang1995@tju.edu.cn}
\credit{Supervision, Methodology, Validation, Writing - Review \& Editing}
\author[2]{Haoran Wang} 
\ead{wanghaoran09@baidu.com}
\credit{Methodology, Writing - Review \& Editing}
\author[1]{Yanwei Pang} 
\ead{pyw@tju.edu.cn}
\credit{Conceptualization, Writing - Review \& Editing}
\author[3]{Xuelong Li} 
\ead{li@nwpu.edu.cn}
\credit{Methodology, Writing - Review \& Editing}

\address[1]{School of Electrical and Information Engineering, Tianjin Key Laboratory of Brain-inspired Intelligence Technology, Tianjin University, Tianjin, 300072, China}
\address[2]{Baidu Research, Beijing, 100193, China}    
\address[3]{School of Artificial Intelligence, OPtics and ElectroNics (iOPEN) and the Key Laboratory of Intelligent Interaction and Applications, Ministry of Industry and Information Technology, Northwestern Polytechnical University, Xi'an, 710072, China}

\cortext[1]{Corresponding author}

\begin{abstract}
	As a promising field, Multi-Query Image Retrieval (MQIR) aims at searching for the semantically relevant image given multiple region-specific text queries. Existing works mainly focus on a single-level similarity between image regions and text queries, which neglects the hierarchical guidance of multi-level similarities and results in incomplete alignments. Besides, the high-level semantic correlations that intrinsically connect different region-query pairs are rarely considered. To address above limitations, we propose a novel Hierarchical Matching and Reasoning Network (HMRN) for MQIR. It disentangles MQIR into three hierarchical semantic representations, which is responsible to capture fine-grained local details, contextual global scopes, and high-level inherent correlations. HMRN comprises two modules: Scalar-based Matching (SM) module and Vector-based Reasoning (VR) module. Specifically, the SM module characterizes the multi-level alignment similarity, which consists of a fine-grained local-level similarity and a context-aware global-level similarity. Afterwards, the VR module is developed to excavate the potential semantic correlations among multiple region-query pairs, which further explores the high-level reasoning similarity. Finally, these three-level similarities are aggregated into a joint similarity space to form the ultimate similarity. Extensive experiments on the benchmark dataset demonstrate that our HMRN substantially surpasses the current state-of-the-art methods. For instance, compared with the existing best method Drill-down, the metric R@1 in the last round is improved by 23.4\%. Our source codes will be released at \textcolor{blue}{https://github.com/LZH-053/HMRN}.
\end{abstract}



\maketitle

\section{Introduction}
\begin{figure}[t]
	\centering
	\includegraphics[width=\columnwidth]{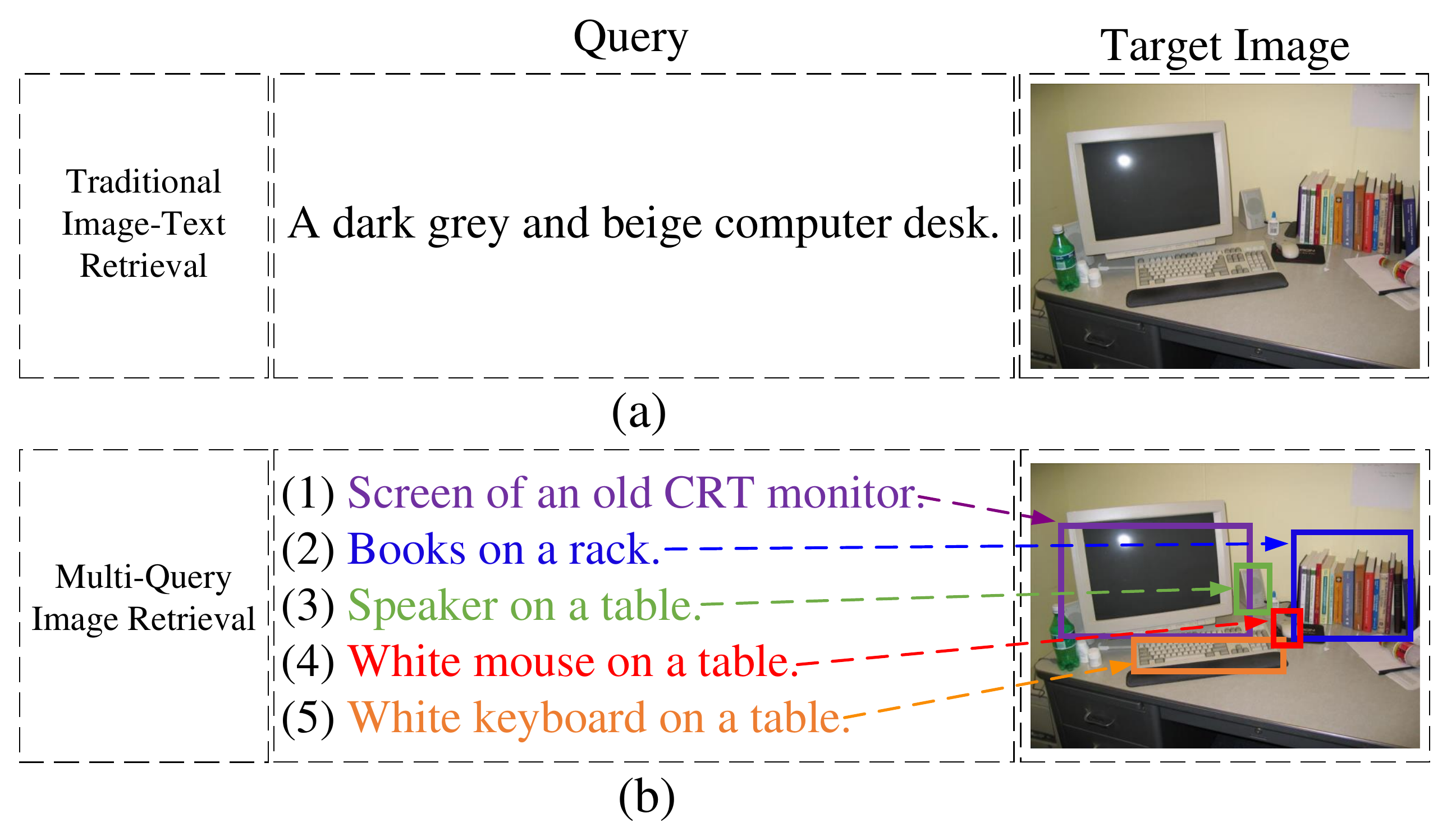}
	\caption{Conceptual comparison of traditional ITR and MQIR. (a) ITR performs image retrieval with a single query, which contains coarse-grained global semantic information, while MQIR performs image retrieval with multiple queries, which contains more fine-grained correspondences.}\label{fig_1}
	\end{figure}

Multi-Query Image Retrieval (MQIR) refers to searching for the semantically relevant image given a set of region-specific queries. As a emerging topic, MQIR provides a more realistic and challenging paradigm than the traditional Image-Text Retrieval (ITR) \citep{scan,GPO,naaf,zhao2023generative,wang2023quaternion}, where the latter retrieves the target image according to a coarse-grained textual query, as shown in Fig. \ref{fig_1}(a). Differently in Fig. \ref{fig_1}(b), MQIR employs multiple detailed descriptions, which focuses on learning more fine-grained correspondences between vision and language. 

Although substantial efforts have been made in retrieval-related tasks, studies on MQIR remain relatively limited \citep{siddiquie2011image,arandjelovic2012multiple,drilldown,wu2021deconfounded}. As a pioneering work, Drill-down \citep{drilldown} is specially designed for MQIR, which encodes multiple textual queries into a set of state vectors and adopts the unidirectional attention mechanism to calculate the similarity. Despite the remarkable performance, MQIR still faces critical challenges. On the one hand, existing models mostly specialize in a single-level similarity, which is insufficient for precise and comprehensive associations. In other words, due to the more complicated correspondences of MQIR, multi-level similarities are essential to measure the semantic correspondences at different levels. Therefore, a hierarchical model is required to explore and assemble different levels of similarities. On the other hand, the inherent correlation among available region-query pairs has not been employed, which contains rich correlation-enhanced semantic information and possesses the potential to further boost the fine-grained alignments. More intuitively, as shown in Fig \ref{fig_1}(b), the red pair (the red box ``mouse'' and its corresponding description) and the orange pair (the orange box ``keyboard'' and its corresponding description) usually appear simultaneously. Thus, their relationship provides more valuable information to understand this scenario. In summary, we conclude the above limitations as: (1) Current methods focus on the single-level similarity while neglecting the hierarchical effectiveness of the multi-level semantic similarities. (2) The inherent correlation that exists in multiple region-query pairs has not been explored, which is a high-level complementation to the hierarchical structure.

In the past decade, many retrieval-related methods have exploited the advancement of hierarchy \citep{SHAN,guo2023hgan}, while the hierarchical structure has not been explored in MQIR. Previous methods either employ the hierarchical structure to facilitate multi-level alignments or explore the high-level semantic correspondence. The former focuses on leveraging the attention mechanism \citep{attention} to calculate the multi-level similarity \citep{scan,SHAN,hu2019multi,chen2020expressing,ji2022asymmetric,jin2022coarse}. However, they are specialized in the coarse alignment between an image and a single global description, while MQIR demands multiple region-specific queries and contains more fine-grained semantic information, which increases the difficulties to accurately align images and queries. In another word, the generalization abiliy of existing multi-level alignment based methods to MQIR remains unproven. The latter tends to utilize the popular Graph Convolutional Network (GCN) \citep{gcn} and Transformer \citep{attention} to excavate cross-modal correlations and assist the high-level reasoning similarity \citep{guo2023hgan,liu2021hit}. These methods mainly consider the intra-modal correlation within each modality \citep{vsrn,he2021cross,VSRN++} and the inter-modal correlation \citep{RACG} between different modalities. Differently in MQIR, the high-level semantic correlations among multiple region-query pairs are rarely considered.
\begin{figure}[t]
\centering
\includegraphics[width=\columnwidth]{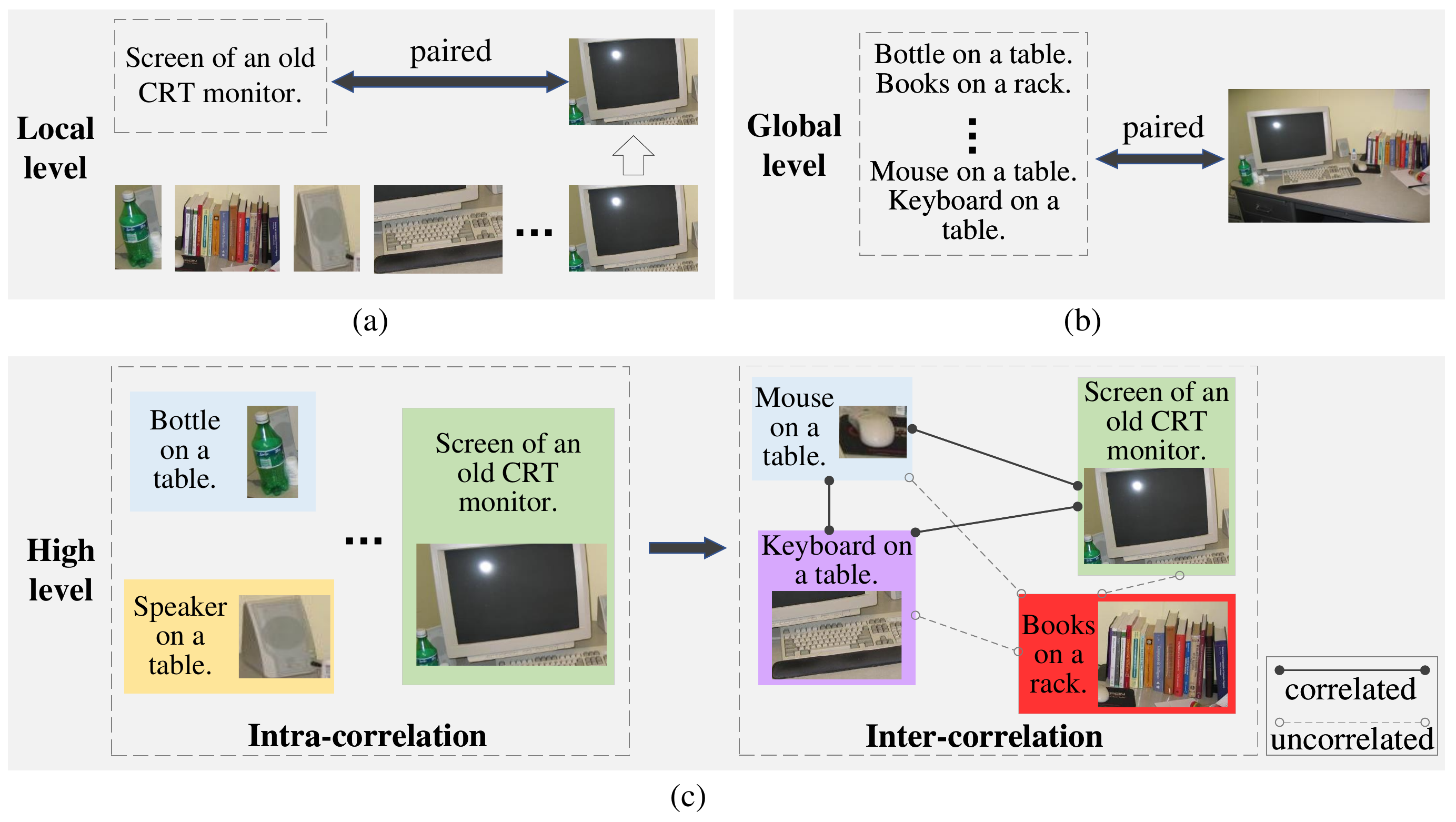}
\caption{Explanation of our structural hierarchy, which disentangles MQIR into three levels. (a) The local level captures the fine-grained alignment between regions and region-specific queries. (b) The global level leverages the context information to conduct the comprehensive alignment. (c) The high level explores the inherent correlations from the perspective of intra-correlations and inter-correlations. The former denotes the semantic correlation within each region-query pair and the latter denotes the semantic correlation among different region-query pairs. }\label{fig_2}
\end{figure}

To address above issues, we propose a Hierarchical Matching and Reasoning Network (HMRN) for MQIR. Depending on different similarity forms, HMRN is composed of two modules: Scalar-based Matching (SM) module and Vector-based Reasoning (VR) module. We disentangle MQIR into a three-level hierarchy: the region-query local level and image-text global level from the SM module, and the correlation-enhanced high-level from the VR module, as shown in Fig. \ref{fig_2}. Specifically, we first leverage the SM module to obtain the multi-level scalar-based similarity considering both the region-query pair from the local level and the image-text pair from the global level. In the SM module, the Local-level Matching approach utilizes the symmetric cross attention mechanism to measure the local-level similarity. The Global-level Matching approach employs the self-attention mechanism to obtain the holistic feature representations and calculates the global-level similarity. Afterwards, we further develop the VR module to explore the vector-based reasoning similarity, where the Intra-Correlation Mining approach is proposed to mine the intra-correlation within each region-query pair, and the Inter-Correlation Reasoning approach aims at reasoning the inter-correlation among different region-query pairs. Finally, we aggregate the three-level similarity of our hierarchical model to form the ultimate similarity, which collaboratively fuses the local-level, global-level, and high-level semantic information.

Our main contributions are summarized as follows:
\begin{itemize}
        \item We propose a novel model that leverages the hierarchical structure to explore diverse characteristics in MQIR, which not only conducts fine-grained alignments, but also captures the high-level semantic correlations. 
        \item We develop a Scalar-based Matching (SM) module to measure the multi-level alignment similarity, which comprises attention-based local-level matching and global-level matching.
        \item To consider the high-level reasoning similarity, we design a Vector-based Reasoning (VR) module to capture the intra-correlation within each region-query pair and the inter-correlation among different region-query pairs, which further enhances the fusion of multi-level similarities at a higher level.
\end{itemize}

The remaining part of this paper is organized as follows. Section 2 introduces the related work. Section 3 elaborates the proposed HMRN model in detail, including five parts: visual and textual representation, scalar-based matching, vector-based reasoning, similarity aggregation strategy, and training objective. Extensive experiments and analyses are presented in Section 4. Finally, Section 5 concludes the paper.

\section{Related Work}
In this part, we extensively surveyed the topics related to our work, which includes image-text retrieval, multi-query image retrieval, and hierarchical structure.
\subsection{Image-text retrieval}
ITR typically employs a single coarse-grained sentence as query to retrieve the semantically relevant image from the image database, and vice versa. Existing methods can be mainly divided into two categories: global correspondence learning methods and local correspondence learning methods.

Global correspondence learning methods aim at mapping the whole image and the full text into holistic embeddings to find their semantic correspondences. Many works focus on improving the global feature encoding network \citep{vsrn,vse++,zheng2020dual} to learn more discriminative representations. For instance, Faghri \emph{et al.} \citep{vse++} utilized a two-stream global feature encoding network and computed the pairwise similarity of global features. Li \emph{et al.} \citep{vsrn,VSRN++} employed the GCN to generate the relationship-enhanced visual and textual representations. Chen \emph{et al.} \citep{GPO} designed an effective generalized pooling operator to aggregate the region-level and word-level features into holistic representations.   

Local correspondence learning methods tend to learn the local alignments between regions and words, which takes local semantic correspondences into account. Many works are devoted to finding precise region-word alignments \citep{scan,ji2019saliency,chen2020imram,zhang2020context,wang2022sum}. Lee \emph{et al.} \citep{scan} proposed one of the most typical methods SCAN, which employs a stacked cross attention framework to focus on the key regions/words and discovers the full latent alignments. Zhang \emph{et al.} \citep{CMCAN} captured the local and global similarities, then calculated the matching confidence between image regions and the complete image, with the text as a bridge. Other works attempt to exploit the cross-modal correlations \citep{vsrn,CMCAN,GSMN,sgraf,zhang2023consensus}. For instance, Diao \emph{et al.} \citep{sgraf} developed a similarity graph reasoning network to infer the relation-aware similarities via capturing the relationships between local and global alignments. Wen \emph{et al.} \citep{wen2021learning} designed a dual semantic relations attention network to enhance the relations between regional objects and global concepts.

\subsection{Multi-query image retrieval}
MQIR describes each image by multiple detailed sentences with richer semantic information, which is more suitable for complicated realistic scenarios. Note that the fine-grained details demanded by MQIR largely increase the difficulties to achieve reliable retrieval. Consequently, it usually necessitates a more delicately designed framework. 

Previous MQIR methods typically treat multiple queries as an integrated set, and collectively input them to retrieve the target image \citep{drilldown,cheng2016effects,han2017learning}. A general solution is to explore the fine-grained alignments between image regions and multiple region-specific queries. Early works \citep{siddiquie2011image,arandjelovic2012multiple} mostly utilize explicit attributes as queries. For example, Siddiquie \emph{et al.} \citep{siddiquie2011image} proposed to integrate multi-label learning and model the correlations among multiple attributes. Arandjelovic \emph{et al.} \citep{arandjelovic2012multiple} built a real-time multi-query retrieval system to retrieve images with specific objects. As the rapid development of the neural network, recent works have enhanced the feature representations to utilize more general text queries instead of attributes or phrases. For instance, Tan \emph{et al.} \citep{drilldown} proposed an effective framework named Drill-down, which leverages regional captions as queries to retrieve the target image. Neculai \emph{et al.} \citep{neculai2022probabilistic} designed a new compositional learning method that composes multi-modal queries into probabilistic embeddings, supporting a combination of multiple query images or texts. However, these methods mainly focus on the single-level similarity \citep{drilldown} and lack a hierarchical structure to aggreagate multi-level similarities. Besides, the semantic correlations existed in potential region-query pairs have not been exploited, which could facilitate the high-level reasoning similarity.

To this end, we not only focus on the multi-level alignment similarity from the local-level and global-level perspectives, but also consider the high-level reasoning similarity from the intra-correlation and inter-correlation perspectives. Finally, three-level similarities are aggregated into the joint similarity space to conduct retrieval.

\subsection{Hierarchical structure}
Hierarchical structure has been successfully applied into many cross-modal fields, such as cross-modal retrieval \citep{ma2022query,yu2022ghan}, visual question answering \citep{yu2018joint,zhang2023spatial}, and visual grounding \citep{li2022hero,tan2023hierarchical}. Rather than merely focusing on a single perspective, hierarchical structure has the potential to explore more sophisticated correspondences. There are several attempts to learn the hierarchical representation for the field of cross-modal retrieval. For instance, Ji \emph{et al.} \citep{SHAN} decomposed image-text retrieval into a multi-step cross-modal matching process, which achieves local-to-local, global-to-local and global-to-global alignments sequentially. Guo \emph{et al.} \citep{guo2023hgan} proposed a hierarchical graph alignment network to construct feature graphs for each modality and enhance the multi-granularity relations between local and global alignments. Other works attempt to exploit the interactions between different levels. Ging \emph{et al.} \citep{ging2020coot} designed a cooperative hierarchical transformer to exploit long-range temporal context based on the interactions between low-level and high-level semantic information. Liu \emph{et al.} \citep{liu2021hit} proposed a hierarchical transformer to exploit the multi-scale features of different transformer layers and perform hierarchical matching with momentum contrast. 

Unlike the above methods, our method designs a customized hierarchical matching and reasoning pattern to aggregate both the multi-level alignment similarity and the high-level reasoning similarity, which captures diverse characteristics for MQIR. 

\section{Methods}

\begin{figure*}[h]
\centering
\includegraphics[width=0.98\linewidth]{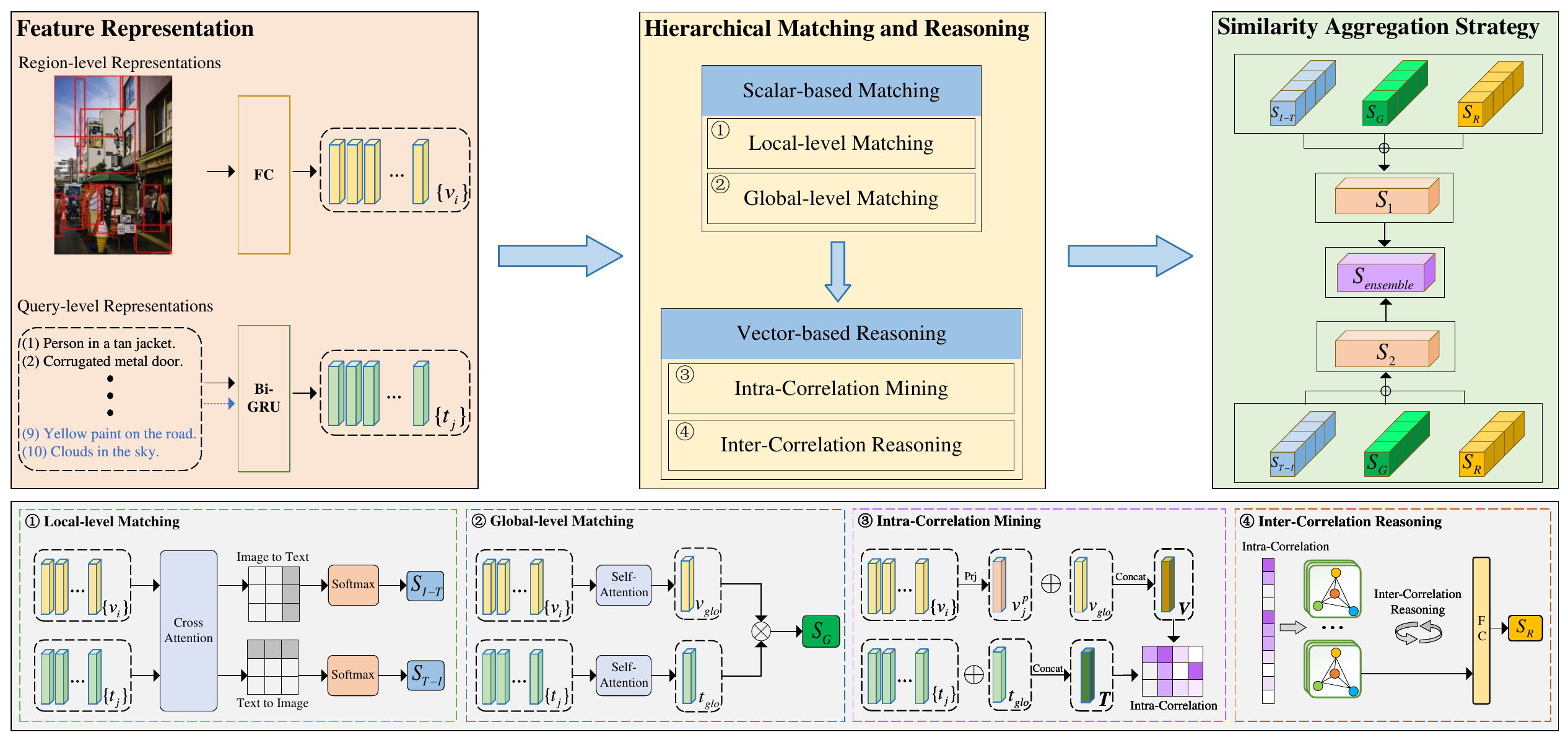}
\caption{An overview of the proposed HMRN method. Firstly, images and queries are encoded into feature representations. Then, depending on different similarity forms, two proposed modules are introduced: \ding{172} The Scalar-based Matching (SM) module focuses on measuring multi-level alignment similarity, including the Local-level Matching approach and the Global-level Matching approach. \ding{173} The Vector-based Reasoning (VR) module attempts to explore the high-level reasoning similarity, which leverages the Intra-Correlation Mining approach to delve the intra-correlation within each region-query pair and employs the Inter-Correlation Reasoning approach to reason the inter-correlation among different region-query pairs. Finally, the three-level similarities are aggregated to form the ultimate similarity. }
\label{fig_3}
\end{figure*}

The overall framework of our proposed Hierarchical Matching and Reasoning Network (HMRN) is shown in Fig. \ref{fig_3}. In the following subsections, we begin with introducing the visual and textual representation. Then, the proposed HMRN model is presented, which consists of two modules: Scalar-based Matching (SM) module and Vector-based Reasoning (VR) module. Specifically, SM aims at obtaining the local-level and global-level similarity, and VR aims at mining the high-level correlations among multiple region-query pairs to obtain the reasoning similarity. Finally, we describe the similarity aggregation strategy as well as the training objective.

\subsection{Visual and textual representation}
To encode visual representations, we take advantage of the bottom-up attention \citep{bottom-up} with the Faster R-CNN\citep{ren2015faster} detector and encode each image as a set of visual embeddings. We select the top $K$ proposals, and represent each detected region with an $X$-dimensional feature $x_i$. Then, a linear projection is employed to transform $x_i$ to a $D$-dimensional embedding $v_i$:
\begin{flalign}\label{equation1}
v_i={\rm\textbf{W}}_v x_i+b_v, & &
\end{flalign}
where ${\rm\textbf{W}}_v \in \mathbb{R}^{X \times D}$ is a learnable parameter matrix and $b_v$ is a bias term. Each image is encoded as ${\rm\textbf{V}}_{loc}=\left\{v_i \mid i \in[1, K], v_i \in \mathbb{R}^D\right\}$, where each $v_i$ represents a specific salient region. 

To encode textual representations, we split each text query into $M$ words and map each word to a one-hot encoding $w_i$, then embed it into a $E$-dimensional embedding $e_i$ via a learnable word embedding layer: $e_i={\rm\textbf{W}}_e w_i$. To obtain the textual representation with semantic context, we utilize the Bi-directional Gated Recurrent Units \citep{bi-gru} (Bi-GRU) to process the word embeddings sequentially and integrate the forward and backward contextual information. The context-aware word representation $h_i$ is obtained by averaging the hidden state of the forward and backward GRU:
\begin{flalign}\label{equation4}
h_i=\frac{\overrightarrow{\operatorname{GRU}}\left(e_i, \overrightarrow{h}_{i-1}\right)+\overleftarrow{\operatorname{GRU}}\left(e_i, \overleftarrow{h}_{i+1}\right)}{2}, i \in[1, M],  & &
\end{flalign}
where $\overrightarrow{\operatorname{GRU}}\left(e_i, \overrightarrow{h}_{i-1}\right)$ and $\overleftarrow{\operatorname{GRU}}\left(e_i, \overleftarrow{h}_{i+1}\right)$ denote the the forward and backward hidden states, respectively. 

We regard the output of the last hidden state as the sentence-level textual representation: $t_j=h_M$. For each target image, $N$-round region-specific queries are provided as ${\rm\textbf{T}}_{loc}=\left\{t_j \mid j \in[1, N], t_j \in \mathbb{R}^D\right\}$, where $t_j$ is the representation of each query. 

\subsection{Scalar-based matching}
Inspired by \citep{scan}, we bridge the heterogeneous gap by developing a multi-level Scalar-based Matching (SM) module, which includes the Local-level Matching approach and the Global-level Matching approach. Particularly, the SM module expects two-level inputs: local representations and global representations. Given these representations, we attend on image regions and text queries differently via a symmetric cross attention mechanism, and calculate the multi-level alignment similarity from the local level and global level.

\subsubsection{Local-level matching}
We conduct local-level matching to measure the fine-grained alignments from two symmetric directions: Image-Text and Text-Image.

\textbf{Image-Text Cross Attention.} We first calculate the cosine similarity for all possible region-query pairs: $s(v_i,t_j)=v_i^T t_j /\left\|v_i\right\|_2\left\|t_j\right\|_2$, where $\|\cdot\|_2$ denotes the $l_2$ normalization. Considering that the region-specific queries have different relevance to each region, we calculate the smilarity $s(v_i,{\rm\textbf{T}}_{loc})$ between each region and different queries with a weight coeffecient $\alpha_{ij}$:
\begin{flalign}\label{equation5}
s(v_i,{\rm\textbf{T}}_{loc})=\sum_{j=1}^N \alpha_{i j} [s(v_i,t_j)]_{+},  & &
\end{flalign}
where
\begin{flalign}\label{equation6}
\alpha_{i j}=\frac{exp \left(\lambda_1 [s(v_i,t_j)]_{+}\right)}{\sum_{j=1}^N exp \left(\lambda_1 [s(v_i,t_j)]_{+}\right)}.  & &
\end{flalign}
The weight coefficient $\alpha_{ij}$ is calculated by the softmax function with a temperature coefficient $\lambda_1$, indicating the relevance of each region with respect to different queries. $[s(v_i,t_j)]_{+}=\max (s(v_i,t_j), 0)$. Finally, the image-text similarity $S_{I-T}$ is obtained as follows:
\begin{flalign}\label{equation7}
S_{I-T}=\frac{1}{K} \sum_{i=1}^K s(v_i,{\rm\textbf{T}}_{loc}).  & &
\end{flalign}

\textbf{Text-Image Cross Attention.} Similarly, we calculate the cosine similarity $s_{j i}$ from another direction. We attend on each query with respect to different image regions, and obtain the similarity $s(t_j,{\rm\textbf{V}}_{loc})$ as follows:
\begin{flalign}\label{equation8}
s(t_j,{\rm\textbf{V}}_{loc})=\sum_{i=1}^K \alpha_{j i} [s(t_j,v_i)]_{+},  & &
\end{flalign}
where
\begin{flalign}\label{equation9}
\alpha_{j i}=\frac{exp \left(\lambda_2 [s(t_j,v_i)]_{+}\right)}{\sum_{i=1}^K exp \left(\lambda_2 [s(t_j,v_i)]_{+}\right)}.  & &
\end{flalign}
The weight coefficient $\alpha_{ji}$ is calculated by the softmax function with a temperature coefficient $\lambda_2$. $[s(t_j,v_i)]_{+}=\max (s(t_j,v_i), 0)$. The text-image similarity is obtained as:
\begin{flalign}\label{equation10}
S_{T-I}=\frac{1}{N} \sum_{j=1}^N s(t_j,{\rm\textbf{V}}_{loc}).  & &
\end{flalign}

\subsubsection{Global-level matching}
Besides the local-level fine-grained alignments, global-level contextual information also plays an important role. We first perform the self-attention mechanism \citep{attention} over all local features to obtain the global representation. For the visual global representation $v_{glo}$, we adopt the average feature $v_{a v e}=\frac{1}{K}\sum_{i=1}^K v_i$ as the simple global feature and aggregate all regions as follows:
\begin{flalign}\label{equation2}
s_i=\frac{exp \left({\rm\textbf{W}}_s\left[\left({\rm\textbf{W}}_{a v e} v_{a v e}\right) \odot\left({\rm\textbf{W}}_r v_i\right)\right]\right)}{\sum_{i=1}^K exp \left({\rm\textbf{W}}_s\left[\left({\rm\textbf{W}}_{a v e} v_{a v e}\right) \odot\left({\rm\textbf{W}}_r v_i\right)\right]\right)},  & &
\end{flalign}
\begin{flalign}\label{equation3}
v_{g l o}=\left\|\sum_{i=1}^K s_i v_i\right\|_2,  & &
\end{flalign}
where ${\rm\textbf{W}}_{ave }$, ${\rm\textbf{W}}_r \in \mathbb{R}^{D \times D}$, ${\rm\textbf{W}}_s \in \mathbb{R}^{D \times 1}$ are learnable parameter matrices. The weight coefficient $s_i$ represents the relevance of $v_i$ and $v_{ave}$, and $\|\cdot\|_2$ denotes the $l_2$ normalization. Similarly, the global textual representation $t_{glo}$ is calculated with the same self-attention mechanism over all text queries.

Afterwards, we conduct global-level matching to measure the global similarity $S_G$ between the self-attended visual global representation $v_{glo}$ and textual global representation $t_{glo}$. The pairwise similarity is measured as:
\begin{flalign}\label{equation11}
S_G=v_{g l o}^T t_{g l o}.  & &
\end{flalign}

\subsection{Vector-based reasoning}
To the best of our knowledge, existing MQIR methods mostly focus on the alignments between images and multiple queries, while neglecting their intrinsical correlations. Note that the intra-correlation within each region-query pair and the inter-correlation among different region-query pairs are able to provide comprehensive semantical guidance. To this end, we delicately design a Vector-based Reasoning (VR) module to explore the semantic correlations existed in multiple region-query pairs. Specifically, we develop the Intra-Correlation Mining approach to investigate the coupled intra-correlation by fusing the visual and textual vectors into a unified representation. Then we employ the Inter-Correlation Reasoning approach to reason the closely linked inter-correlation among different region-query pairs. Finally, based on above inherent correlations, we obtain the high-level reasoning similarity between images and queries.

\subsubsection{Intra-correlation mining}
To lay the foundation for the subsequent Inter-Correlation Reasoning, we first project the visual representation into the same dimensionality as the textual representation, then we form the intra-correlation vector from these two representations. Similar to Eq. (\ref{equation9}), we first apply the Text-Image cross attention:
\begin{flalign}\label{equation12}
\alpha_{j i}^p=\frac{exp \left(\lambda_2 \widetilde{s}{(t_j,v_i)}\right)}{\sum_{i=1}^K exp \left(\lambda_2 \widetilde{s}{(t_j,v_i)}\right)},  & &
\end{flalign}
where $\alpha_{ji}^p$ is the weight coefficient, and $\lambda_2$ is the temperature coefficient of the softmax function. $s(t_j,v_i)$ is normalized by $\widetilde{s}{(t_j,v_i)}=[{s(t_j,v_i)}]_{+} / \sqrt{\sum_{j=1}^N[{s(t_j,v_i)}]_{+}^2}$, and $[{s(t_j,v_i)}]_{+}=\max (s(t_j,v_i), 0)$.
Further, we employ the weight coefficient $\alpha_{ji}^p$ to project the visual local representation ${\rm\textbf{V}}_{loc}$ as follows:
\begin{flalign}\label{equation13}
v_j^p=\sum_{i=1}^K \alpha_{j i}^p v_i,  & &
\end{flalign}
and obtain the transformed visual local representation ${\rm\textbf{V}}_{loc }^p=\left\{v_j^p\mid j \in[1,N], v_j^p \in \mathbb{R}^D\right\}$.

\begin{figure}[t!]
        \centering
        \includegraphics[width=\columnwidth]{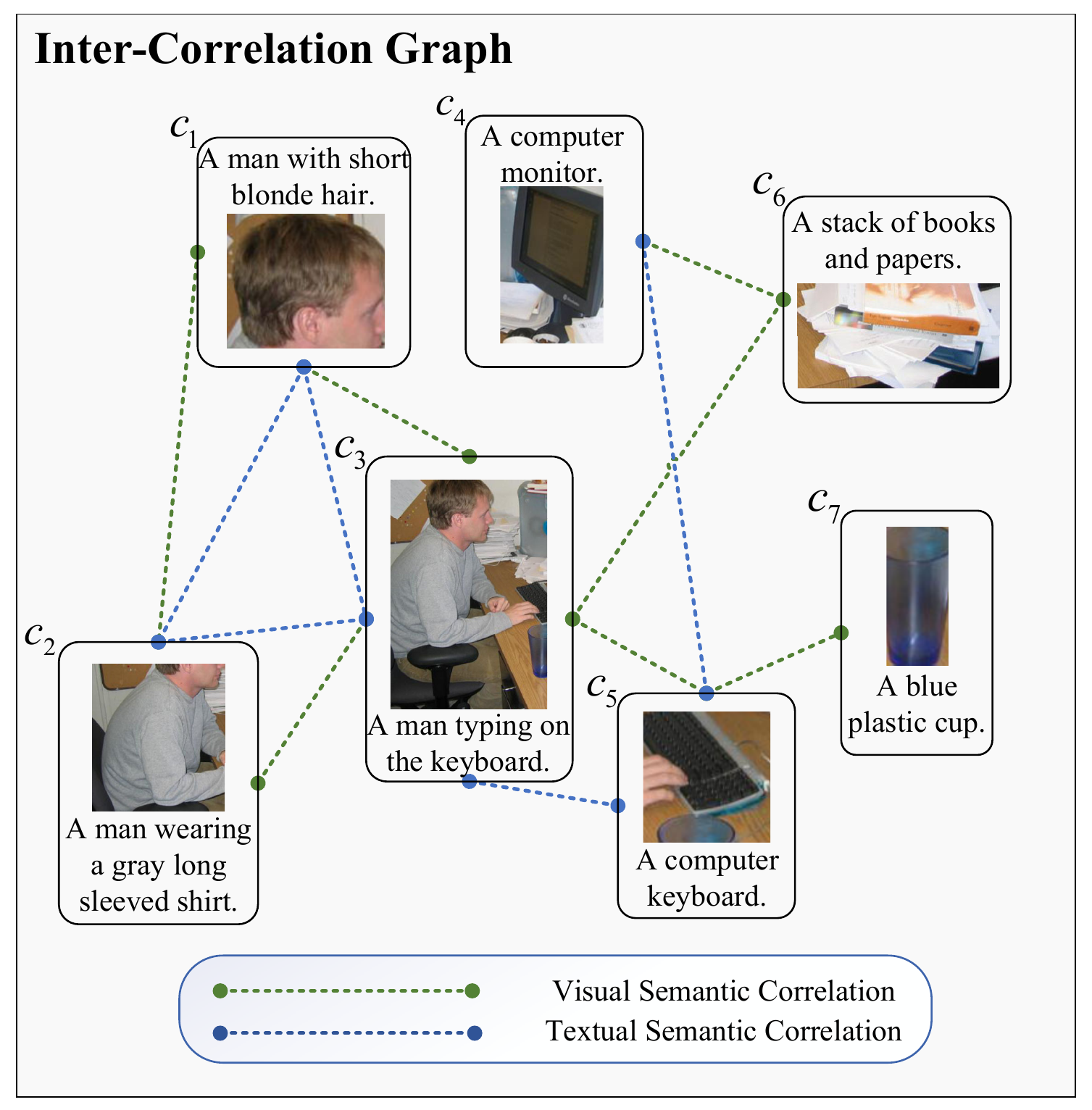}
        \caption{An illustration of the inter-correlation graph among region-query pairs, in which the different colored lines denote different types of semantic correlations.}
        \label{fig_4}
        \end{figure}

Given the projected visual local representation ${\rm\textbf{V}}_{loc}^p$, we build an integrated visual vector ${\rm\textbf{V}}=\left\{v_1^p, \ldots, v_N^p, v_{g l o}\right\}$ by concatenating ${\rm\textbf{V}}_{loc}^p$ and the global visual representation $v_{glo}$. Likewise, the integrated textual vector ${\rm\textbf{T}}=\left\{t_1, \ldots, t_N, t_{g l o}\right\}$ is obtained as well. Then, the overall intra-correlation vector ${\rm\textbf{C}}=\left\{c_1, \ldots, c_N, c_{g l o}\right\}$ is calculated as:
\begin{flalign}\label{equation14}
{\rm\textbf{C}}({\rm\textbf{V}}, {\rm\textbf{T}})=\frac{{\rm\textbf{W}}_c|{\rm\textbf{V}}-{\rm\textbf{T}}|^2}{\left\|{\rm\textbf{W}}_c|{\rm\textbf{V}}-{\rm\textbf{T}}|^2\right\|_2},  & &
\end{flalign}
where $c_j$ represents the intra-correlation within each region-query pair, and ${\rm\textbf{W}}_c \in \mathbb{R}^{D \times D}$ is a learnable parameter matrix.

\subsubsection{Inter-correlation reasoning}
Besides the coupled intra-correlation, the inter-correlation among different region-query pairs is also vital to characterize the relation-aware semantic information. As shown in Fig. \ref{fig_4}, the inter-correlation can be empirically summarized as the visual semantic correlation and the textual semantic correlation. Concretely, the visual semantic correlation refers to the overlap or semantic associations of different regions, such as the ``grey shirt'' in nodes $c_2$ and $c_3$, and the textual semantic correlation refers to the co-occurred contextual words or phrases, such as ``keyboard'' and ``computer keyboard'' in nodes $c_3$ and $c_5$, respectively.

We build a vector-based inter-correlation graph to explore the semantic inter-correlation among different region-query pairs. Inspired by \citep{sgraf}, the local region-query pairs and the global image-text pair consisting in the intra-correlation vector ${\rm\textbf{C}}$ are constructed as the graph nodes. The edge sets between graph nodes are calculated by an affinity matrix {\rm\textbf{A}}:
\begin{flalign}\label{equation15}
{\rm\textbf{A}}\left({\rm\textbf{C}}_q, {\rm\textbf{C}}_k\right)=\frac{exp \left(\left({\rm\textbf{W}}_q {\rm\textbf{C}}_q\right)^T\left({\rm\textbf{W}}_k {\rm\textbf{C}}_k\right)\right)}{\sum_q exp \left(\left({\rm\textbf{W}}_q {\rm\textbf{C}}_q\right)^T\left({\rm\textbf{W}}_k {\rm\textbf{C}}_k\right)\right)},  & &
\end{flalign}
where ${\rm\textbf{W}}_q, {\rm\textbf{W}}_k \in \mathbb{R}^{D \times D}$ are learnable parameter matrices for ingoing query nodes and outgoing key nodes, respectively. Eq. (\ref{equation15}) shows that the affinity matrix {\rm\textbf{A}} is obtained by measuring the edge of each adjacent node. Hence, the edges with strong inter-correlations are given higher affinity scores.

As described above, the nodes of the inter-correlation graph are constructed by the intra-correlation vector within each region-query pair. We consider these vector-based nodes are competent to deliver the high-level semantic correlations. Response of each graph node is obtained by considering the inter-correlation to its adjacent nodes. We update the nodes as follows:
\begin{flalign}\label{equation16}
{\rm\textbf{C}}^{l+1}=\operatorname{ReLU}\left({\rm\textbf{W}}_R \sum_q {\rm\textbf{A}}\left({\rm\textbf{C}}_q^l, {\rm\textbf{C}}_k^l\right) \cdot {\rm\textbf{C}}_q^l\right),  & &
\end{flalign}
where ${\rm\textbf{W}}_R \in \mathbb{R}^{D \times D}$ is a learnable parameter matrix to map the attended graph nodes to the next step, and ReLU denotes the Rectified Linear Unit.

We implement the inter-correlation graph reasoning for $L$ steps. The final graph nodes set ${\rm\textbf{C}}^*=\left\{c_1^*, \ldots ,c_N^*, c_{g l o}^*\right\}$ is regarded as the semantic correlation-enhanced representation. We take the output of the global node $c_{glo}^*$ as the final high-level semantic representation, and feed it into a fully-connected layer with the ReLU activation function to obtain the reasoning similarity $S_R$. 

In summary, the VR module exploits the high-level intra-correlation and inter-correlation existed in multiple region-query pairs, which further complements the hierarchical structure and facilitates the multi-level alignments.

\subsection{Similarity aggregation strategy}
To excavate the full potential of our model, we divide our model into two individual models, namely HMRN I-T and HMRN T-I, and train them separately. HMRN I-T contains local-level similarity $S_{I-T}$, global-level similarity $S_G$, and high-level reasoning similarity $S_R$, while HMRN T-I contains another local-level similarity $S_{T-I}$, along with the other two levels of similarity. We employ a similarity aggregation strategy to fuse the multi-level similarity into the ultimate similarity. Specifically, We first calculate the similarity of each individual model. Then, we obtain the ensemble similarity $S_{ensemble}$ by summing the similarity of the two individual models:
\begin{flalign}\label{equation19}
S_{ensemble}=S_1 + S_2,  & &
\end{flalign}
\begin{flalign}\label{equation20}
S_1=\alpha S_{I-T}+\beta S_R+(1-\alpha-\beta) S_G,  & &
\end{flalign}
\begin{flalign}\label{equation21}
S_2=\alpha S_{T-I}+\beta S_R+(1-\alpha-\beta) S_G,  & &
\end{flalign}
where $\alpha$ and $\beta$ are weight factors. $S_1$, $S_2$ denotes the aggregated similarity of HMRN I-T and HMRN T-I, respectively.

\subsection{Training objective}
To enlarge the distance between the matched image-text pairs and the mismatched pairs, we leverage the popular InfoNCE loss \citep{infonce} as our training objective, which has been employed in many retrieval related tasks \citep{dong2022partially,Han2021TextBasedPS}. With the ground-truth image-text pair $(v_g,t_g)$, the InfoNCE loss is calculated as:
\begin{equation}\label{equation17}
\begin{split}
&\mathcal{L}_{nce}=-\frac{1}{n} \sum_{(v, t) \in \mathcal{B}}\left[log \left(\frac{exp \left(\tau \cdot S(v_g, t_g)\right)}{\sum_{i}^n exp \left(\tau \cdot S\left(v_g, t_i\right)\right)}\right)\right. \\   
&\left.+log \left(\frac{exp \left(\tau \cdot S(v_g, t_g)\right)}{\sum_{i}^n exp \left(\tau \cdot S\left(v_i, t_g\right)\right)}\right)\right], & 
\end{split}
\end{equation}
where $S(\cdot)$ denotes the similarity between image and text. $n$ denotes the number of samples in a mini-batch, and $\tau$ is a temperature coeffecient.

We employ the InfoNCE loss on the three-level hierarchical similarities. Note that we train the individual model HMRN I-T and HMRN T-I separately. Therefore, the training loss for each model is calculated as:
\begin{flalign}\label{equation18}
\mathcal{L}=\alpha \mathcal{L}_{S_{L}}+\beta \mathcal{L}_{S_R}+(1-\alpha-\beta) \mathcal{L}_{S_G}, & & 
\end{flalign}
where $S_L$ denotes the local-level similarity $S_{I-T}$ for HMRN I-T and $S_{T-I}$ for HMRN T-I. $S_R$ denotes the high-level reasoning similarity, and $S_G$ denotes the global-level similarity. $\alpha$ and $\beta$ balance the weight of different losses, which are identical with those in Eq. (\ref{equation20}) and Eq. (\ref{equation21}).

\begin{table*}[t!]\centering
        \caption{Comparisons of image retrieval results on R@1. Note that * indicates re-implementing and revising open-source codes for MQIR.\label{tab:table1}} 
        \resizebox{0.9\linewidth}{!}{
        \begin{tabular}{cccccccccccc}
        \hline
        \multirow{2}{*}{Paradigms}                        & \multirow{2}{*}{Methods} & \multicolumn{10}{c}{R@1 for each round}                                       \\
        &                          & 1             & 2             & 3             & 4             & 5             & 6             & 7             & 8             & 9             & 10            \\ \hline
        \multirow{3}{*}{\begin{tabular}[c]{@{}c@{}} \makecell{Single \\ -Vector} \end{tabular}} & R-RankFusion$_{\rm{NeurIPS'19}}$ \citep{drilldown}      & 6.5           & 12.2          & 15.2          & 16.2          & 17.0          & 17.6          & 17.8          & 18.2          & 18.1          & 18.3          \\
        & R-RE$_{\rm{NeurIPS'19}}$ \citep{drilldown}                     & 4.9           & 9.7           & 14.5          & 17.6          & 21.5          & 24.8          & 27.9          & 30.2          & 32.2          & 33.9          \\
        & R-HRE$_{\rm{NeurIPS'19}}$ \citep{drilldown}                    & 4.9           & 9.5           & 14.6          & 18.9          & 23.5          & 27.4          & 31.1          & 34.1          & 36.3          & 38.7          \\ \hline
        \multirow{9}{*}{\begin{tabular}[c]{@{}c@{}} \makecell{Multi \\ -Vector} \end{tabular}}  & Drill-down$_{\rm{NeurIPS'19}}$ \citep{drilldown}              & 5.7           & 18.7          & 31.5          & 41.7          & 49.2          & 55.9          & 60.7          & 64.6          & 67.5          & 70.1          \\
        & VSRN$^*_{\rm{ICCV'19}}$ \citep{vsrn}         & 5.7           & 14.1          & 22.4          & 29.0          & 35.0          & 40.3          & 44.6          & 48.4          & 51.5          & 53.9          \\
        & SGR$^*_{\rm{AAAI'21}}$ \citep{sgraf}         & 3.2           & 13.3          & 22.9          & 31.1          & 39.2          & 45.5          & 50.5          & 55.6          & 58.9          & 62.1          \\
        & SAF$^*_{\rm{AAAI'21}}$ \citep{sgraf}      & 5.1           & 14.8          & 24.3          & 32.6          & 40.6          & 45.6          & 51.6          & 56.6          & 60.1          & 63.2          \\
        & SGRAF$^*_{\rm{AAAI'21}}$ \citep{sgraf}       & 5.3           & 15.5          & 25.8          & 34.9          & 43.0            & 49.7          & 54.9          & 59.9          & 63.1          & 66.1          \\
        & NAAF$^*_{\rm{CVPR'22}}$ \citep{naaf}    & 4.9           & 16.3          & 27.4          & 36.6          & 44.3          & 50.4          & 55.4          & 60.1          & 63.0            & 65.9          \\
        & HMRN T-I (ours)          & 6.8           & 20.7          & 33.7          & 44.1          & 52.6          & 59.4          & 65.1          & 69.8          & 73.2          & 75.7          \\
        & HMRN I-T (ours)          & 18.5          & 33.8          & 49.1          & 62.0            & 72.9          & 80.9          & 85.9          & 89.3          & 91.5          & 93.0            \\
        & HMRN ensemble (ours)     & \textbf{36.8} & \textbf{58.7} & \textbf{71.7} & \textbf{78.9} & \textbf{83.8} & \textbf{87.6} & \textbf{89.7} & \textbf{91.4} & \textbf{92.4} & \textbf{93.5} \\ \hline
        \end{tabular}}
        \end{table*}

\section{Experiments}
\subsection{Datasets and evaluation metrics}
We evaluate our proposed method on the benchmark Visual Genome dataset \citep{vg}, in which each image is labeled with multiple region-specific captions. Duplicate region captions and images with less than 10 captions are removed. Following \citep{drilldown}, the remaining 105,414 images are further split into 92,105/5,000/9,896 for training/validation/testing. Note that the testing split is not used for the training of the object detector \citep{bottom-up} which is employed in our image encoding branch. Queries and their orders are randomly sampled during training, while during validation and testing, they are kept fixed.

We adopt the popular Recall at K (R@K, K=1, 5, 10) and Mean Rank (MR) as the evaluation metrics. R@K indicates the percentage of target images in the retrieved top K ranking list. MR indicates the average ranks of all target images. We evaluate the overall performance over all rounds by averaging the above metrics, which is defined as Average Recall at K (Avg. R@K, K=1, 5, 10) and Average Mean Rank (Avg. MR), respectively. Besides, Average R@Sum (Avg. R@Sum) denotes the summation of all Avg. R@K metrics.  All the evaluations are performed on the test split. 

\subsection{Implementation details}
Following \citep{drilldown}, we train the model with 10-round ($N$=10) queries. For each image, we utilize the Faster-RCNN\citep{ren2015faster} detector provided by \citep{bottom-up} with ResNet-101\citep{resnet} as the backbone to encode the top $K$=36 region proposals and obtain an $X$=2048 dimensional feature for each region. For each query, we set the dimensionality of the word embedding $E$ as 300 and the number of hidden states in Bi-GRU as 256. The dimensionality $D$ of the similarity representation is set to 256. In VR module, the Inter-Correlation Reasoning approach iterates $L$ = 3 steps. For other hyperparameters, we set temperature coefficients $\lambda_1$, $\lambda_2$, and $\tau$ as 5, 15, and 40, respectively. Our model is trained by the Adam optimizer \citep{adam} with the mini-batch size of 128. All models are trained for 150 epochs with the learning rate of $4e^{-4}$ for the first 75 epochs and $4e^{-5}$ for the next 75 epochs. We select the model with the best performance on the validation set for testing.

\begin{table*}[t!]\centering
\caption{Comparisons of image retrieval results on R@5. Note that * indicates re-implementing and revising open-source codes for MQIR.\label{tab:table2}} 
\resizebox{0.9\linewidth}{!}{
\begin{tabular}{cccccccccccc}
\hline
\multirow{2}{*}{Paradigms}                                                              & \multirow{2}{*}{Methods} & \multicolumn{10}{c}{R@5 for each round}                                                                                                                       \\
&                          & 1             & 2             & 3             & 4             & 5             & 6             & 7             & 8             & 9             & 10            \\ \hline
\multirow{3}{*}{\begin{tabular}[c]{@{}c@{}}\makecell{Single \\ -Vector} \end{tabular}} & R-RankFusion$_{\rm{NeurIPS'19}}$ \citep{drilldown}             & 17.9          & 26.2          & 29.1          & 30.4          & 31.2          & 31.4          & 31.2          & 31.8          & 32.3          & 32.7          \\
& R-RE$_{\rm{NeurIPS'19}}$ \citep{drilldown}                     & 14.3          & 25.6          & 34.3          & 40.9          & 46.2          & 50.6          & 55.3          & 58.3          & 61.3          & 63.4          \\
& R-HRE$_{\rm{NeurIPS'19}}$ \citep{drilldown}                    & 14.5          & 26.1          & 35.7          & 42.7          & 48.4          & 53.9          & 58.1          & 61.7          & 64.9          & 67.1          \\ \hline
\multirow{9}{*}{\begin{tabular}[c]{@{}c@{}}\makecell{Multi \\ -Vector} \end{tabular}}  & Drill-down$_{\rm{NeurIPS'19}}$ \citep{drilldown}               & 17.2          & 38.9          & 54.8          & 64.9          & 71.7          & 77.3          & 80.8          & 83.7          & 85.8          & 87.5          \\
& VSRN$^*_{\rm{ICCV'19}}$ \citep{vsrn}                     & 16.6          & 33.1          & 44.6          & 53.5          & 60.2          & 65.3          & 70.1          & 73.6          & 76.1          & 78.5          \\
& SGR$^*_{\rm{AAAI'21}}$ \citep{sgraf}                      & 12.6          & 31.1          & 45.4          & 55.9          & 63.8          & 69.9          & 74.4          & 78.9          & 81.4          & 83.4          \\
& SAF$^*_{\rm{AAAI'21}}$ \citep{sgraf}                      & 14.5          & 33.5          & 47.1          & 57.1          & 65.1          & 70.5          & 75.0          & 79.3          & 81.8          & 83.9          \\
& SGRAF$^*_{\rm{AAAI'21}}$ \citep{sgraf}                    & 14.9          & 34.6          & 49.4          & 59.7          & 67.8          & 73.0          & 77.5          & 81.1          & 83.9          & 85.6          \\
& NAAF$^*_{\rm{CVPR'22}}$ \citep{naaf}                    & 14.6          & 35.0          & 49.9          & 60.1          & 66.9          & 73.1          & 76.7          & 80.5          & 83.2          & 85.0          \\
& HMRN T-I (ours)          & 18.7          & 41.6          & 58.1          & 67.8          & 75.4          & 80.3          & 84.1          & 86.9          & 89.2          & 90.9          \\
& HMRN I-T (ours)          & 37.9          & 57.6          & 72.8          & 82.6          & 89.3          & 93.4          & 95.8          & 97.1          & 97.6          & \textbf{98.3}          \\
& HMRN ensemble (ours)     & \textbf{60.6} & \textbf{79.5} & \textbf{88.1} & \textbf{91.9} & \textbf{94.3} & \textbf{95.7} & \textbf{96.8} & \textbf{97.3} & \textbf{97.7} & 98.1 \\ \hline
\end{tabular}}
\end{table*}

\begin{table*}[t]\centering
\caption{Comparisons of image retrieval results on R@10. Note that * indicates re-implementing and revising open-source codes for MQIR.\label{tab:table3}}
\resizebox{0.9\linewidth}{!}{ 
\begin{tabular}{cccccccccccc}
\hline
\multirow{2}{*}{Paradigms}                                                              & \multirow{2}{*}{Methods} & \multicolumn{10}{c}{R@10 for each round}                                                                                                                      \\
&                          & 1             & 2             & 3             & 4             & 5             & 6             & 7             & 8             & 9             & 10            \\ \hline
\multirow{3}{*}{\begin{tabular}[c]{@{}c@{}}\makecell{Single \\ -Vector} \end{tabular}} & R-RankFusion$_{\rm{NeurIPS'19}}$ \citep{drilldown}             & 25.8          & 33.5          & 36.5          & 38.0            & 38.3          & 38.9          & 39.1          & 39.5          & 40.1          & 40.3          \\
& R-RE$_{\rm{NeurIPS'19}}$ \citep{drilldown}                     & 21.4          & 35.8          & 45.6          & 53.0            & 58.7          & 63.1          & 67.1          & 70.4          & 73.3          & 74.9          \\
& R-HRE$_{\rm{NeurIPS'19}}$ \citep{drilldown}                    & 22.2          & 35.9          & 47.3          & 55.1          & 61.4          & 66.4          & 70.6          & 73.7          & 76.1          & 78.6          \\ \hline
\multirow{9}{*}{\begin{tabular}[c]{@{}c@{}}\makecell{Multi \\ -Vector} \end{tabular}}  & Drill-down$_{\rm{NeurIPS'19}}$ \citep{drilldown}               & 25.4          & 49.7          & 64.6          & 73.3          & 79.7          & 83.9          & 86.8          & 88.9          & 90.8          & 92.0            \\
& VSRN$^*_{\rm{ICCV'19}}$ \citep{vsrn}                     & 24.0            & 43.5          & 55.3          & 64.1          & 70.3          & 75.3          & 78.6          & 81.9          & 83.9          & 85.8          \\
& SGR$^*_{\rm{AAAI'21}}$ \citep{sgraf}                      & 19.6          & 42.0            & 57.0            & 66.4          & 73.3          & 78.3          & 82.0            & 85.4          & 87.8          & 89.3          \\
& SAF$^*_{\rm{AAAI'21}}$ \citep{sgraf}                      & 22.0            & 43.6          & 57.6          & 67.2          & 74.1          & 79.2          & 82.7          & 86.0            & 88.1          & 89.7          \\
& SGRAF$^*_{\rm{AAAI'21}}$ \citep{sgraf}                    & 22.7          & 45.3          & 60.1          & 69.6          & 78.4          & 81.0            & 84.4          & 87.2          & 89.3          & 91.0            \\
& NAAF$^*_{\rm{CVPR'22}}$ \citep{naaf}                    & 22.2          & 44.9          & 60.2          & 69.1          & 75.9          & 80.6          & 83.7          & 86.5          & 88.6          & 90.0            \\
& HMRN T-I (ours)          & 27.5          & 52.4          & 67.7          & 76.3          & 82.4          & 86.5          & 89.6          & 91.3          & 93.2          & 94.5          \\
& HMRN I-T (ours)          & 46.5          & 66.6          & 79.9          & 87.8          & 92.7          & 95.7          & 97.3          & 98.3          & \textbf{98.7}          & \textbf{99.0}            \\
& HMRN ensemble (ours)     & \textbf{68.4} & \textbf{84.8} & \textbf{91.8} & \textbf{94.4} & \textbf{96.2} & \textbf{97.3} & \textbf{97.8} & \textbf{98.8} & 98.5 & 98.8 \\ \hline
\end{tabular}}
\end{table*}

\subsection{Performance comparison and analysis}
We compare our proposed HMRN with the recent state-of-the-art methods on the Visual Genome \citep{vg} dataset. Depending on the number of vectors that the model encodes the query features into, existing methods can be categorized as single-vector based methods and multi-vector based methods.

We first follow Drill-down \citep{drilldown} to introduce several single-vector based methods. Particularly, we choose three methods: R-RankFusion \citep{drilldown}, R-RE \citep{drilldown}, and R-HRE \citep{drilldown}. (a) R-RankFusion calculates the ranks for each round separately, and the final results are obtained by averaging the ranks from each round. (b) R-RE is a recurrent encoder network and utilizes a uni-directional GRU to encode the concatenation of queries. (c) R-HRE employs a hierarchical text encoder to encode the queries, which is similar to recent dialog based methods \citep{guo2018dialog,liao2018knowledge}. It obtains textual representations by taking the current query and previous hidden features into account. Since the above three methods represent the queries as a single vector, we classify them as single-vector based methods.

Afterwards, we introduce the up to date multi-vector based methods. As methods specifically designed for MQIR are rare, we reproduce the traditional ITR methods for further comparison. We select several open-source methods and re-implement their official codes to be consistent with the setting of MQIR, including VSRN \citep{vsrn}, SGRAF \citep{sgraf}, and NAAF \citep{naaf}. Concretely, we leverage these traditional ITR methods to calculate the similarity between images and each round of query. Then, we average these similarities by $N$-round for ranking. We retrain these comparative methods on the Visual Genome \citep{vg} dataset for fair and reproducible comparison. These methods are grouped into multi-vector based methods since they obtain the query representation as multiple vectors. Note that our HMRN and the up-to-date MQIR method Drill-down \citep{drilldown} fall in this group.

Tables \ref{tab:table1} to \ref{tab:table4} present the results of R@1, R@5, R@10, and MR, respectively. We also show the results of Avg. R@K, Avg. R@Sum and Avg. MR in Table \ref{tab:table5} to assess the overall performance. We report the results of the individual model HMRN T-I, HMRN I-T and the HMRN ensemble model, respectively. Based on these results, we have several observations and conclusions as below.

\begin{table*}[t]\centering
\caption{Comparisons of image retrieval results on MR. Note that * indicates re-implementing and revising open-source codes for MQIR.\label{tab:table4}}
\resizebox{0.9\linewidth}{!}{
\begin{tabular}{cccccccccccc}
\hline
\multirow{2}{*}{Paradigms}                                                              & \multirow{2}{*}{Methods} & \multicolumn{10}{c}{MR for each round}                                                                                                               \\
&                          & 1            & 2             & 3             & 4             & 5            & 6            & 7            & 8          & 9            & 10           \\ \hline
\multirow{3}{*}{\begin{tabular}[c]{@{}c@{}}\makecell{Single \\ -Vector} \end{tabular}} & R-RankFusion$_{\rm{NeurIPS'19}}$ \citep{drilldown}             & 657.9        & 521.7         & 431.3         & 386.7         & 356.0          & 337.8        & 322.2        & 313.6      & 305.2        & 296.6        \\
& R-RE$_{\rm{NeurIPS'19}}$ \citep{drilldown}                     & 697.2        & 224.1         & 121.1         & 84.7          & 63.2         & 47.4         & 35.7         & 27.6       & 23.1         & 19.8         \\
& R-HRE$_{\rm{NeurIPS'19}}$ \citep{drilldown}                    & 565.8        & 186.8         & 102.2         & 73.9          & 54.1         & 41.3         & 30.6         & 23.8       & 19.4         & 16.4         \\ \hline
\multirow{9}{*}{\begin{tabular}[c]{@{}c@{}}\makecell{Multi \\ -Vector} \end{tabular}}  & Drill-down$_{\rm{NeurIPS'19}}$ \citep{drilldown}               & 321.8        & 119.6         & 59.1          & 37.6          & 25.3         & 18.7         & 13.9         & 10.4       & 8.3          & 6.6          \\
& VSRN$^*_{\rm{ICCV'19}}$ \citep{vsrn}                     & 434.1        & 189.3         & 107.8         & 74.3          & 54.0           & 41.1         & 31.4         & 23.2       & 18.1         & 14.9         \\
& SGR$^*_{\rm{AAAI'21}}$ \citep{sgraf}                       & 343.9        & 152.9         & 83.1          & 53.4          & 36.2         & 26.1         & 19.7         & 14.9       & 11.7         & 9.2          \\
& SAF$^*_{\rm{AAAI'21}}$ \citep{sgraf}                      & 335.4        & 148.8         & 78.2          & 51.6          & 35.4         & 25.2         & 18.6         & 14.0         & 10.8         & 8.7          \\
& SGRAF$^*_{\rm{AAAI'21}}$ \citep{sgraf}                     & 319.6        & 137.4         & 71.8          & 46.5          & 31.4         & 22.5         & 16.7         & 12.5       & 9.7          & 7.7          \\
& NAAF$^*_{\rm{CVPR'22}}$ \citep{naaf}                    & 367.5        & 148.1         & 75.9          & 49.1          & 33.2         & 24.4         & 18.3         & 13.6       & 10.5         & 8.3          \\
& HMRN T-I (ours)          & 249.6        & 94.1          & 45.9          & 29.8          & 20.5         & 15.0           & 10.3         & 7.6        & 5.6          & 4.4          \\
& HMRN I-T (ours)          & 192.2        & 72.6          & 37.1          & 21.4          & 13.4         & 8.5          & 4.7          & 3.1        & \textbf{2.3} & \textbf{1.8} \\
& HMRN ensemble (ours)     & \textbf{136.0} & \textbf{40.7} & \textbf{19.4} & \textbf{11.9} & \textbf{8.6} & \textbf{6.5} & \textbf{4.2} & \textbf{3.0} & \textbf{2.3} & 2.0            \\ \hline
\end{tabular}}
\end{table*}

\begin{table*}[t]
\caption{Comparisons of image retrieval results on Avg. R@K(K=1, 5, 10), Avg. R@Sum, and Avg. MR. Note that * indicates re-implementing and revising open-source codes for MQIR.\label{tab:table5}}
\renewcommand\arraystretch{1.1}  
\begin{tabular*}{0.9\textwidth}{@{}@{\extracolsep{\fill}}ccccccc@{}} 
\hline
\multirow{2}{*}{Paradigms}                                                              & \multirow{2}{*}{Methods} & \multicolumn{5}{c}{Average results of 10 rounds}                                               \\
&                          & R@1      & R@5      & R@10     & \multicolumn{1}{c}{R@Sum} & MR       \\ \hline
\multirow{3}{*}{\begin{tabular}[c]{@{}l@{}}\makecell{Single \\ -Vector} \end{tabular}} & R-RankFusion$_{\rm{NeurIPS'19}}$ \citep{drilldown}             & 15.7          & 29.4          & 37.0          & 82.1                           & 392.9         \\
& R-RE$_{\rm{NeurIPS'19}}$ \citep{drilldown}                     & 21.7          & 45.0          & 56.3          & 123.0                          & 134.4         \\
& R-HRE$_{\rm{NeurIPS'19}}$ \citep{drilldown}                    & 23.9          & 47.3          & 58.7          & 129.9                          & 111.4         \\ \hline
\multirow{9}{*}{\begin{tabular}[c]{@{}l@{}}\makecell{Multi \\ -Vector} \end{tabular}}  & Drill-down$_{\rm{NeurIPS'19}}$ \citep{drilldown}                & 46.6          & 66.3          & 73.5          & 186.4                          & 62.1          \\
& VSRN$^*_{\rm{ICCV'19}}$ \citep{vsrn}                     & 34.5          & 57.2          & 66.3          & 158.0                          & 98.8          \\
& SGR$^*_{\rm{AAAI'21}}$ \citep{sgraf}                      & 38.2          & 59.7          & 68.1          & 166.0                           & 75.1          \\
& SAF$^*_{\rm{AAAI'21}}$   \citep{sgraf}                    & 39.5          & 60.8          & 69.0          & 169.3                          & 72.7          \\
& SGRAF$^*_{\rm{AAAI'21}}$  \citep{sgraf}                   & 41.8          & 62.8          & 70.9          & 175.5                          & 67.6          \\
& NAAF$^*_{\rm{CVPR'22}}$ \citep{naaf}                     & 42.4          & 62.5          & 70.2          & 175.1                          & 74.9          \\
& HMRN T-I (ours)          & 50.1          & 69.3          & 76.1          & 195.5                          & 48.3          \\
& HMRN I-T (ours)          & 67.7          & 82.2          & 86.3          & 236.2                          & 35.7          \\
& HMRN ensemble(ours)      & \textbf{78.5} & \textbf{90.0} & \textbf{92.7} & \textbf{261.2}                 & \textbf{23.5} \\ \hline
\end{tabular*}
\end{table*}

\subsubsection{Comparison with state-of-the-art methods} As the number of queries increases, the performance progressively improves, and our proposed HMRN model surpasses all competitive methods with a clear margin. The highest R@1 is 93.5\%, achieved by HMRN ensemble in the 10th round. On R@5 and R@10, our model also achieves notable gains. Compared with Drill-down \citep{drilldown}, the existing best method, HMRN ensemble significantly improves 31.1\% on R@1 in the first round, and 23.4\% in the last round. As for MR, HMRN ensemble improves 185.8 (57.7\%) in the first round and 4.6 (70.0\%) in the last round. In Table \ref{tab:table5}, we provide a comprehensive comparison by observing the results of Avg. R@K, Avg. R@Sum and Avg. MR, which largely outperforms those of Drill-down by 31.9\%, 23.7\%, 19.2\%, 74.8\%, and 38.6 (62.2\%), respectively. To sum up, the consistently outstanding performance of HMRN demonstrates its effectiveness.    

\subsubsection{Comparison of individual and ensemble model} As shown in Table \ref{tab:table5}, HMRN I-T achieves more advanced performance than HMRN T-I on all metrics, which outperforms the latter by 17.6\%, 12.9\%, and 10.2\% on Avg. R@K, 40.7\% on Avg. R@Sum, and 12.6 (26.1\%) on Avg. MR. The comparison results indicate that the I-T branch contains more cross-modal correspondences to distinguish matched image-text pairs. We argue the reason is that images intuitively contain more context information than texts. Therefore, many meaningful image regions may be overlapping, resulting the corresponding textual descriptions partially similar. In this case, the I-T cross attention allows the model to carefully seek the correct query for each region from the query database, which greatly improves the retrieval accuracy and robustness. Afterwards, we compare the individual models with the HMRN ensemble model. As illustrated in Tables \ref{tab:table1} to \ref{tab:table4}, we observe that the HMRN ensemble model surpasses the HMRN individual model in most rounds. However, when providing more queries, especially in Rounds 9 and 10, the performance of the ensemble model will be degraded by the weaker T-I branch since the I-T branch already provides eminently precise local-level correspondences. As shown in Table \ref{tab:table5}, the HMRN ensemble model surpasses the HMRN individual models on Avg. R@K, Avg. R@Sum, and Avg. MR as well.

\subsubsection{Comparison of different vector-based methods} From Tables \ref{tab:table1} to \ref{tab:table4}, we observe that existing single-vector based methods and multi-vector based methods are on the same level in the first two rounds. However, after the third round, the latter surpasses the former by an increasing margin. The reasons stem from that single vector is insufficient to convey gradually increasing semantic information. Besides, the single-vector query representation may forget the information contained in the previous queries, while multi-vector based methods encode each query separately, thus preserving all the textual semantic information.

\subsubsection{Comparison with traditional retrieval methods} As shown in Tables \ref{tab:table1} to \ref{tab:table5}, we observe that Drill-down \citep{drilldown} and our HMRN outperform the traditional ITR methods. We summarize the reasons into two aspects. On the one hand, traditional methods are specifically developed for relatively simple scenes, such as MSCOCO \citep{MSCOCO} and Flickr30K \citep{Flickr30K}, which could not conduct fine-grained alignments in complex realistic scenarios. On the other hand, traditional retrieval methods are more suitable for scenarios where images are described with a single coarse-grained description, which lacks the abililty to handle multiple region-specific queries and explore the high-level correlations among them.

\subsection{Ablation study}
In this section, we perform a series of ablation studies to validate the effectiveness of the hierarchical structure and the impact of each component.

\begin{table}[t]\centering
\caption{Ablation study on the hierarchical framework of HMRN. ``$S_{I-T}$'' and $S_{T-I}$ denotes the local-level similarity. ``$S_G$'' denotes the global-level similarity, and ``$S_R$'' denotes the high-level reasoning similarity. \label{tab:table6}}
\setlength\tabcolsep{3pt}
\resizebox{\columnwidth}{!}{
\renewcommand\arraystretch{1.1}  
\begin{tabular}{c|cccc|ccccc}
\hline
\multirow{2}{*}{Methods} & \multirow{2}{*}{$S_{I-T}$}   & \multirow{2}{*}{$S_{T-I}$}   & \multirow{2}{*}{$S_G$}     & \multirow{2}{*}{$S_R$}     & \multicolumn{5}{c}{Average results of 10 rounds}                      \\
&                           &                           &                           &                           & R@1           & R@5           & R@10          & R@Sum & MR            \\ \hline
1                        & \checkmark                 &                           &                           &                           & 22.3          & 44.8          & 55.9          & 123.0 & 65.1          \\
2                        &                           &                           & \checkmark &                           & 31.0          & 53.8          & 63.1          & 147.9 & 90.7          \\
3                        &                           &                           &                           & \checkmark & 43.7          & 66.7          & 68.8          & 179.2 & 97.7          \\
4                        &                           & \checkmark &                           &                           & 47.3          & 67.3          & 74.5          & 189.1 & 49.9          \\
5                        &                           & \checkmark & \checkmark &                           & 48.2          & 68.2          & 75.3          & 191.7 & 48.6          \\
6                        & \checkmark &                           & \checkmark &                           & 58.3          & 78.1           & 83.5          & 219.9 & 32.9          \\
7                 &                           & \checkmark & \checkmark & \checkmark & 50.1          & 69.3          & 76.1          & 195.5 & 48.3          \\
8                 & \checkmark &                           & \checkmark & \checkmark & 67.7          & 82.2          & 86.3          & 236.2 & 35.7          \\
9            & \checkmark & \checkmark & \checkmark & \checkmark & \textbf{78.5} & \textbf{90.0} & \textbf{92.7} & \textbf{261.2} & \textbf{23.5} \\ \hline
\end{tabular}}
\end{table}

\subsubsection{Configurations of the hierarchical structure}
We provide different configurations of our hierarchical structure to verify its effectiveness. The results of ablation studies are shown in Table \ref{tab:table6}. From the evaluation results, we draw the following observations and conclusions.
\begin{itemize}
        \item  In Methods 1 to 4, we first investigate the performance of our model along with each single-level similarity, which reflects the base performance before the hierarchical aggregation. Method 4 shows that the T-I attended local-level similarity achieves the best performance, obtaining 189.1 in R@Sum. Comparing Method 1 with 4, the T-I cross attention shows better performance than the I-T branch. We argue this is the reason why most MQIR methods simply leverage the T-I cross attention mechanism, such as the pioneering work Drill-down \citep{drilldown}. By contrast, Method 2 shows relatively inferior performance due to the fact that global-level similarity lacks fine-grained semantic information. Besides, Method 3 studies the effectiveness of the reasoning similarity and results in moderate accuracy.
        \item  Subsequently, we conduct a series of ablation studies on multi-level similarity aggregation to investigate the effect of hierarchy. Methods 5 and 6 incorporate the global-level similarity with the local-level similarity and achieve better results, which proves that the global-level similarity complements the contextual information lacked by the local level and the local-level similarity provides the fine-grained correspondences lacked by the global level. Methods 7 and 8 further combine the reasoning similarity to the multi-level alignment similarities and yields around 1.9\% and 9.4\% improvements on Avg. R@1, respectively, which proves that exploring the high-level semantic correlations from region-query pairs is beneficial to facilitate the retrieval process. Moreover, comparing Methods 7, 8 with Method 3, the remarkable improvements demonstrate that single high-level reasoning similarity is also insufficient and requires multi-level alignments to lay the semantic foundation then discovers its full potential. Finally, we fuse all similarities and obtain the best performance of the ensemble model in Method 9. In summary, it is evident that our HMRN brings consistent benefits by aggreagating multi-level similarities hierarchically.
\end{itemize}
   
\begin{table}[t]\centering
\caption{The impact of different Intra-Correlation Mining modes. ``Concat'' denotes vector concatenation and  ``Sub'' denotes vector subtraction.\label{tab:table7}}
\setlength\tabcolsep{3pt}
\resizebox{\columnwidth}{!}{
\renewcommand\arraystretch{1.2}  
\begin{tabular}{c|cccc|ccccc}
\hline
\multirow{2}{*}{Methods} & \multirow{2}{*}{I-T} & \multirow{2}{*}{T-I} & \multicolumn{2}{c|}{Mode} & \multicolumn{5}{c}{Average results of 10 rounds}              \\
&                      &                      & Concat        & Sub       & R@1       & R@5       & R@10   & R@Sum   & MR        \\ \hline
1                        &                      & \checkmark                    & \checkmark             &           & 49.7 & 69.0          & 75.7   &  194.4      & 48.8          \\
2                        &                      & \checkmark                    &               & \checkmark         & \textbf{50.1} & \textbf{69.3} & \textbf{76.1} & \textbf{195.5} & \textbf{48.3} \\ \hline
3                        & \checkmark                    &                      & \checkmark             &           & \textbf{67.7} & 82.1 & 86.2     & 236.0   & 37.2          \\
4                        & \checkmark                    &                      &               & \checkmark         & \textbf{67.7} & \textbf{82.2} & \textbf{86.3} & \textbf{236.2} & \textbf{35.7} \\ \hline
\end{tabular}}
\end{table}

\subsubsection{Configurations of the VR module}
We investigate the impact of different Intra-Correlation Mining modes in Table \ref{tab:table7}. ``Concatenation'' means directly concatenating visual vectors and textual vectors, and ``Subtraction'' means calculating the difference between visual and textual vectors as Eq. (\ref{equation14}). From the results, we observe that ``Subtraction'' is superior than ``Concatenation'', which mines more intrinsical associations of two different modalities.

\begin{table}[t]\centering
\caption{The impact of different Inter-Correlation Reasoning configurations. Note that ``ICR Step'' denotes the number of steps for Inter-Correlation Reasoning.\label{tab:table8}}
\setlength\tabcolsep{3pt}
\resizebox{\columnwidth}{!}{
\renewcommand\arraystretch{1.1}  
\begin{tabular}{c|cccc|ccccc}
        \hline
        \multirow{2}{*}{Methods} & \multicolumn{4}{c|}{ICR Step} & \multicolumn{5}{c}{Average results of 10 rounds}                               \\
                                 & 1        & 2        & 3        & 4        & R@1           & R@5           & R@10          & R@Sum          & MR            \\ \hline
        1                        & \checkmark        &          &          &          & 78.4          & 89.9          & 92.5          & 260.8          & 24.0          \\
        2                        &          & \checkmark         &          &          & 78.2          & 89.7          & 92.4          & 260.3          & 24.9          \\
        3                        &          &          & \checkmark         &          & \textbf{78.5} & \textbf{90.0} & \textbf{92.7} & \textbf{261.2} & \textbf{23.5} \\
        4                        &          &          &          & \checkmark         & 78.3          & 89.8          & 92.5          & 260.6          & 24.5          \\ \hline
        \end{tabular}}
\end{table}

Table \ref{tab:table8} shows the impact of different Inter-Correlation Reasoning configurations. We conduct experiments on HMRN ensemble to investigate the optimal Inter-Correlation Reasoning step. As we can see, the best performance is obtained when iterating the Inter-Correlation Reasoning approach for three times, which exceeds other settings on all five evaluation metrics. 

\begin{table}[t]\centering
\caption{The impact of training strategies. ``Joint'' and ``Split'' denote joint training and separate training for the I-T and T-I cross attention mechanism, respectively.\label{tab:table9}}
\setlength\tabcolsep{2pt}
\resizebox{\columnwidth}{!}{
\renewcommand\arraystretch{1.1}  
\begin{tabular}{c|cccc|ccccc}
\hline
\multirow{2}{*}{Methods} & \multirow{2}{*}{I-T} & \multirow{2}{*}{T-I} & \multirow{2}{*}{Joint} & \multirow{2}{*}{Split} & \multicolumn{5}{c}{Average results of 10 rounds}              \\
&                      &                      &                        &                        & R@1       & R@5       & R@10  & R@Sum    & MR        \\ \hline
1                        & \checkmark                    & \checkmark                    & \checkmark                      &                        & 75.4        & 87.9          & 91.0  &  254.3      & 25.9 \\
2                        & \checkmark                    & \checkmark                    &                        & \checkmark                      & \textbf{78.5} & \textbf{90.0} & \textbf{92.7} & \textbf{261.2} & \textbf{23.5}          \\ \hline
\end{tabular}}
\end{table}

\subsubsection{Configurations of training strategies} 
We report the results of different training strategies in Table \ref{tab:table9}. Comparing Method 1 with Method 2, we observe that training models separately and assembling them as Eq. \ref{equation19} gains higher performance. We argue that separate training tends to discover greater potential of the cross attention mechanism. By training separately, the I-T and T-I cross attention mechanism align images and texts in two directions thoroughly, without being affected by each other. Then, the two individual model are integrated to reach superior retrieval performance. As for the joint training, it optimizes the symmetric cross attention mechanism simultaneously, thus  trading precision for computational efficiency.

\begin{table}[t]\centering
\caption{The impact of different textual backbones.\label{tab:table10}}
\setlength\tabcolsep{4pt}
\resizebox{\columnwidth}{!}{
\renewcommand\arraystretch{1.1}  
\begin{tabular}{c|ccccc}
\hline
\multirow{2}{*}{Textual Backbone} & \multicolumn{5}{c}{Average results of 10 rounds}              \\
& R@1      &R@5      & R@10 & R@Sum    & MR       \\ \hline
CLIP (ViT/B-16)                &  78.1         & 89.5          &  92.3  & 259.9       & 27.1          \\
Bi-GRU                         & \textbf{78.5} & \textbf{90.0} & \textbf{92.7} & \textbf{261.2} & \textbf{23.5} \\ \hline
\end{tabular}}
\end{table}

\subsubsection{Configurations of the textual backbone} 
Recently, the pre-trained multi-modal models such as CLIP \citep{clip}, ALIGN \citep{align}, and UNITER \citep{uniter} have significantly boosted the performance on a broad range of downstream vision-language tasks. To evaluate the performance of pre-trained models in MQIR, we employ the pre-trained CLIP (ViT/B-16) model as the textual backbone and apply a fully-connected layer to project the CLIP textual features into the same $D$ dimensional representations. The rest settings are kept unchanged. In Table \ref{tab:table10}, we observe that adopting the pre-trained CLIP model slightly degrades the performance. The reasons may stem from two aspects. On the one hand, the Bi-GRU based model typically contains 14,284 words for Visual Genome dataset, while CLIP contains 49,152 words \citep{clip}. In consequence, the latter may be more applicable to various applications but less specialized. On the other hand, due to limited computing resources, we can not finetune the CLIP based textual backbone on the Visual Genome dataset, which may not discover its full potential. Still and all, we observe that the CLIP based model converges much faster than the Bi-GRU based model during the training stage, which saves computational costs and embodies the superiority of the pre-trained models.

\subsection{Sensitivity analysis of hyperparameters}
\begin{figure}[t]
        \centering
        \includegraphics[width=\columnwidth]{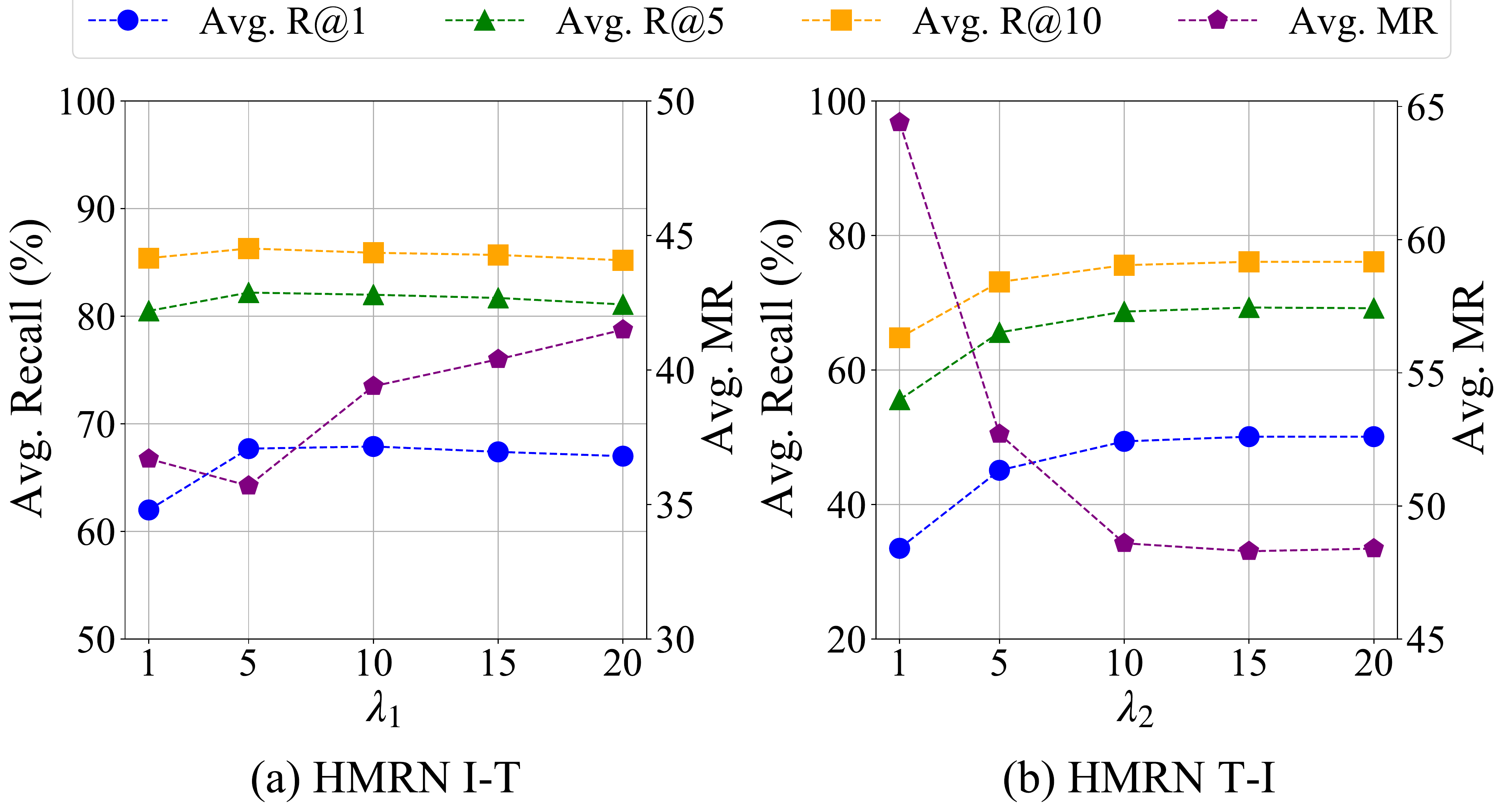}
        \caption{Impacts of $\lambda$ for HMRN I-T and HMRN T-I. Note that Avg. Recall refers to the left vertical coordinate and Avg. MR refers to the right vertical coordinate.}
        \label{fig_5}
        \end{figure}

\begin{figure}[t]
        \centering
        \includegraphics[width=\columnwidth]{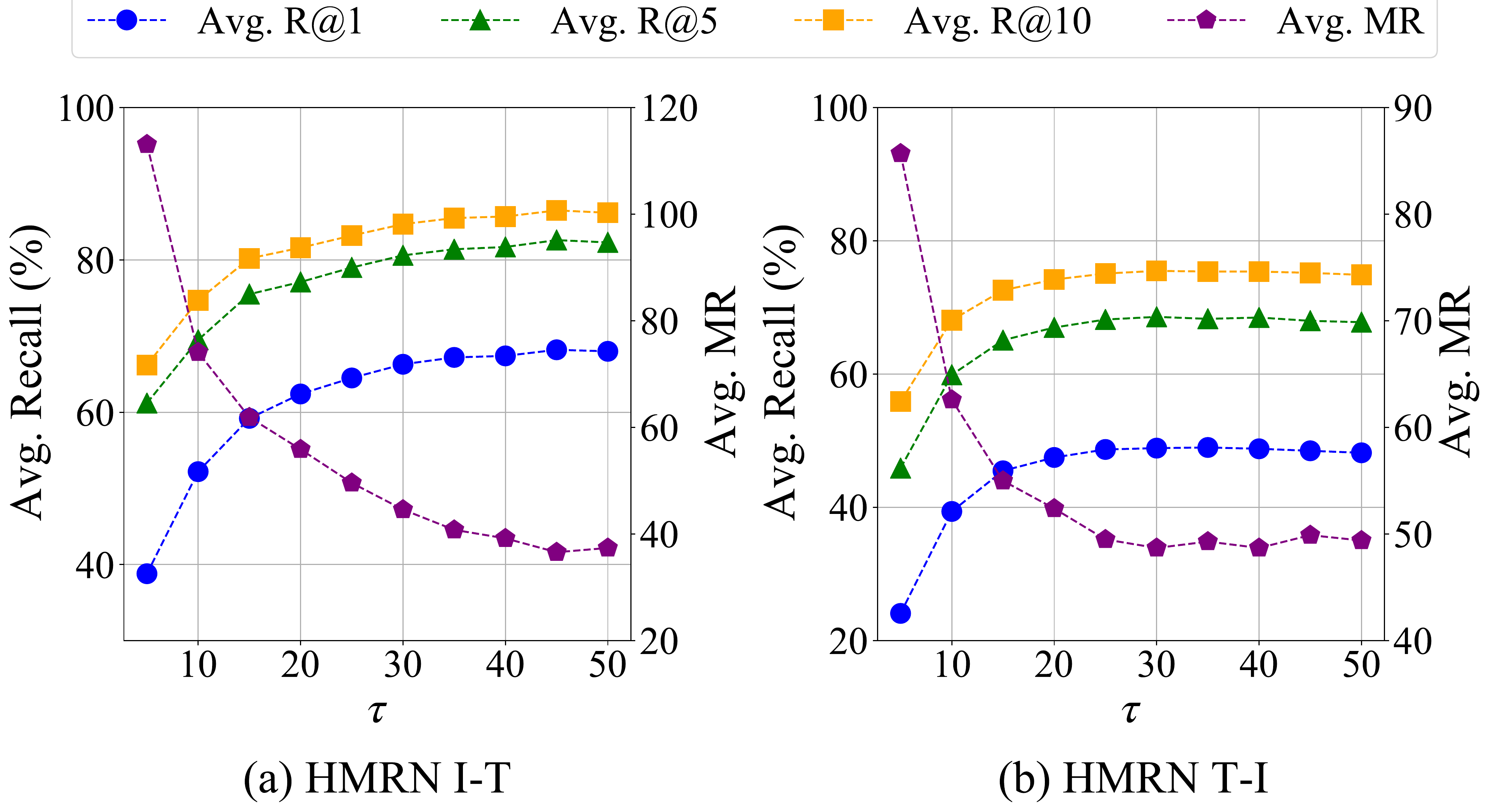}
        \caption{Impacts of $\tau$ for HMRN I-T and HMRN T-I. Note that Avg. Recall refers to the left vertical coordinate and Avg. MR refers to the right vertical coordinate.}
        \label{fig_6}
        \end{figure}
\begin{figure}[t]
        \centering
        \includegraphics[width=\columnwidth]{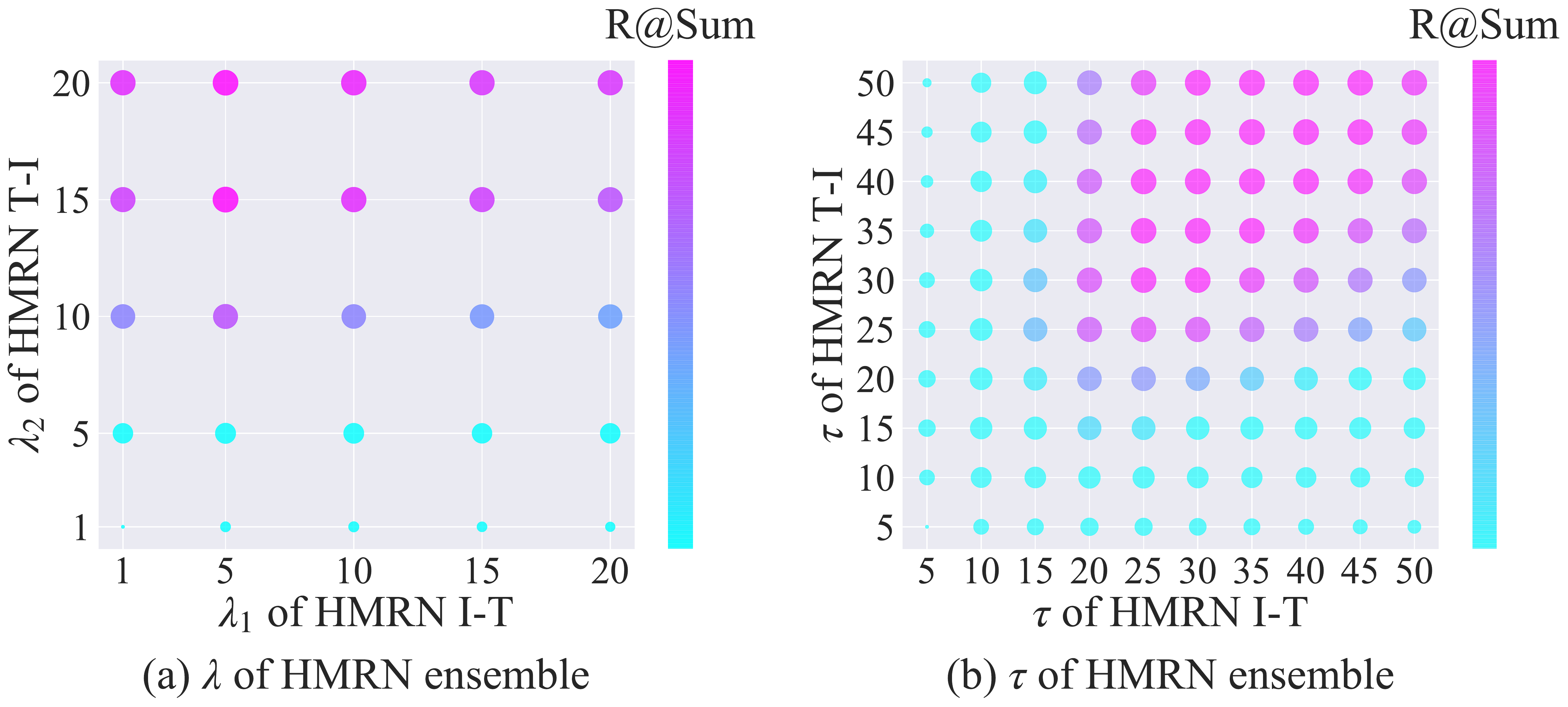}
        \caption{Impacts of $\lambda$ and $\tau$ for HMRN ensemble. Bubbles with warmer color and larger size indicate higher R@Sum.}
        \label{fig_7}
        \end{figure}
\begin{figure}[t]
        \centering
        \includegraphics[width=0.4\textwidth,height=0.32\textwidth]{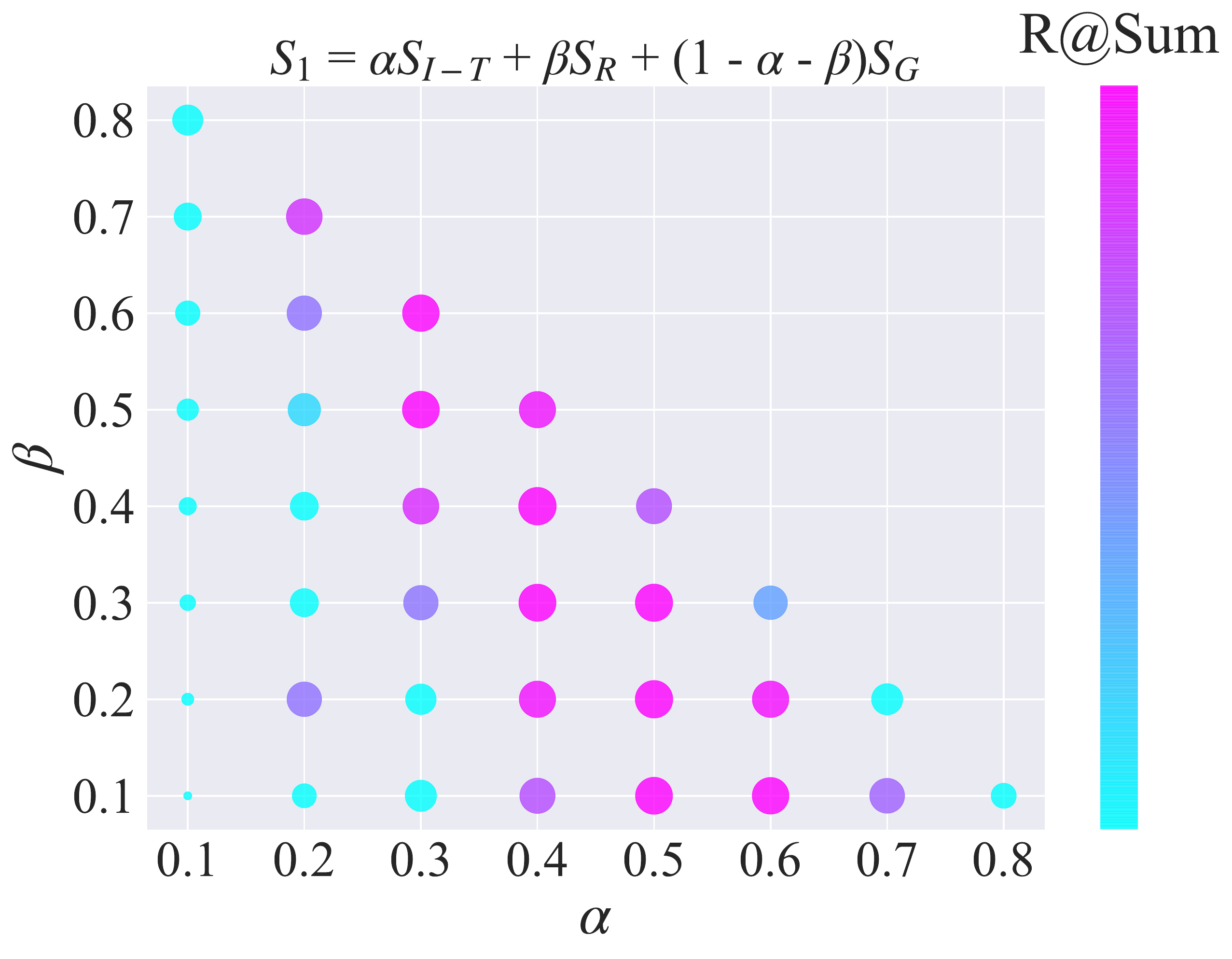}
        \caption{Impacts of similarity weight factors $\alpha$ and $\beta$ on HMRN I-T. Bubbles with warmer color and larger size indicate higher R@Sum. Note that HMRN I-T and HMRN T-I have the same similarity components. For the sake of brevity, we choose one branch HMRN I-T to exemplify the impact of weight factors.}
        \label{fig_8}
        \end{figure}
Our model draws on several hyperparameters, namely $\lambda$, $\tau$, $\alpha$, and $\beta$. $\lambda$ and $\tau$ are temperature coefficients (TempCoef) that control the degree of emphasizing on the matched pairs and punishing the mismatched pairs, while $\alpha$ and $\beta$ are weight factors for multi-level similarity aggregation. Considering that proper selection of these hyperparameters is essential to achieve high performance, we perform exhaustive experiments to explore their sensitivity.

\subsubsection{Analysis of the cross attention TempCoef}
As shown in Fig. \ref{fig_5} (a) and (b), as $\lambda$ grows, the model performance first increases then decreases. We argue the reason is that $\lambda$ controls the degree of punishment for negative samples \citep{wang2021understanding}. Therefore, small $\lambda$ may impede the model's discriminative ability, while large $\lambda$ punishes the negative samples too much and may be counter-productive. Note that above analysis is conducted on the individual models HMRN T-I and HMRN I-T. To study the effect of temperature coefficient $\lambda$ for the HMRN ensemble model, we integrate the similarity as Eq. (\ref{equation19}), and employ Avg. R@Sum to represent the retrieval results of different parameter combinations. Fig. \ref{fig_7} (a) shows that the best Avg. R@Sum is achieved when $\lambda$ is set to 5 and 15 for HMRN I-T and HMRN T-I, respectively. In light of the fact that the HMRN ensemble model obtains stronger performance than the HMRN individual models, We set 5, 15 for $\lambda_1$ and $\lambda_2$ as the final combination.

\subsubsection{Analysis of the InfoNCE TempCoef}
As shown in Eq. (\ref{equation17}), the InfoNCE temperature coefficient $\tau$ highlights the logits of positive image-text pairs to distinguish them. From Fig. \ref{fig_6} (a) and (b), we could observe that as the InfoNCE temperature coefficient $\tau$ increases, the retrieval performance first increments rapidly, then fluctuates within a narrow range. Similar to $\lambda$, we integrate the similarity of the HMRN individual models to study the effect of $\tau$ for the HMRN ensemble model. As shown in Fig. \ref{fig_7} (b), the best Avg. R@Sum is acquired when the temperature coefficient $\tau$ is set to 40 for both HMRN I-T and HMRN T-I. Therefore, we set the $\tau$ to 40 as the final InfoNCE temperature coefficient.

\subsubsection{Analysis of the similarity weight factor}
To collaboratively aggregate the hierarchical three-level similarities, we experimentally conduct studies on the impact of similarity weight factors. $\alpha$, $\beta$ represents the weight of the local-level similarity and the high-level reasoning similarity, respectively. Then the weight of the global-level similarity is calculated as $1-\alpha-\beta$. As shown in Fig. \ref{fig_8}, we have found that our model exhibits competitive R@Sum when $\alpha$ ranges from 0.4 to 0.6 and $\beta$ ranges from 0.2 to 0.4, while the best R@Sum is achieved when $\alpha$ and $\beta$ are both set to 0.4. At this point, the corresponding weight factor of global-level similarity is 0.2. 

Above experiments on similarity weight factors provide further evidence to support the rationale behind the hierarchical structure of our model, which aggregates the multi-level alignment similarity and the high-level reasoning similarity hierarchically to obtain the ultimate similarity. Extensive experiments substantiate that incorporating different levels of similarities allows for a comprehensive understanding of visual and textual modalities.

\subsection{Analysis on computation complexity}
Besides the retrieval performance, computation complexity is another essential evaluation metric. Complex models may provide satisfactory performance, but the huge computation cost hinders their deployment in practice. Therefore, we also evaluate the computation complexity for the proposed HMRN model and make comparisons with the state-of-the-art methods.

\begin{table}[t]\centering
\caption{Comparisons of the computation complexity with the state-of-the-art methods.  Note that * indicates re-implementing and revising open-source codes for MQIR.} \label{tab:table11}
\normalsize
\setlength{\tabcolsep}{3pt}
\resizebox{\columnwidth}{!}{
\renewcommand\arraystretch{1.1}  
\begin{tabular}{c|cc|cccc}
\hline
\multirow{2}{*}{Methods} & \multirow{2}{*}{Image Backbone} & \multirow{2}{*}{Text Backbone} & \multirow{2}{*}{Param(M)} & \multirow{2}{*}{GFLOPs} & \multicolumn{1}{l}{\multirow{2}{*}{Avg. R@1}} \\
&              &                                &                           &                         & \multicolumn{1}{l}{}                         \\ \hline
Drill-down$_{\rm{NeurIPS'19}}$ \citep{drilldown}             & Faster-RCNN                     & GRU                            & \textbf{3.1}             & 2.4       & 46.6                         \\ 
VSRN$^*_{\rm{ICCV'19}}$ \citep{vsrn}              & Faster-RCNN                     & Bi-GRU                            & 111.0                    & 10.4                   & 34.5                \\
SGR$^*_{\rm{SGRAF'21}}$ \citep{sgraf}               & Faster-RCNN                     & Bi-GRU                         & 15.6                     & 2.3                    & 38.2               \\
SAF$^*_{\rm{SGRAF'21}}$ \citep{sgraf}              & Faster-RCNN                     & Bi-GRU                         & 15.0                     & 2.3                    & 39.5                 \\
SGRAF$^*_{\rm{SGRAF'21}}$ \citep{sgraf}      & Faster-RCNN                     & Bi-GRU                         & 30.5                     & 4.6                    & 41.8              \\
NAAF$^*_{\rm{CVPR'22}}$ \citep{naaf}                   & Faster-RCNN                     & Bi-GRU                         & 10.3                     & 1.6                    & 42.4            \\
HMRN ensemble (ours)           & Faster-RCNN                     & Bi-GRU                         & 4.7                   & \textbf{0.5}        & \textbf{78.5}                     \\ \hline
\end{tabular}}
\end{table}

Table \ref{tab:table11} presents the experimental results on computation complexity. On the one hand, we observe that with the same image backbone Faster-RCNN \citep{ren2015faster} and similar GRU-based text backbone, our model has 4.7M trainable parameters, which is much lighter than most counterparts. In comparison with the current best method Drill-down \citep{drilldown}, HMRN is slightly heavier but outperforms it more than 30\%  on Avg. R@1. On the other hand, with the input batch size of 1, HMRN achieves 78.5\% Avg. R@1 while only requiring 0.5 GFLOPs. Despite the distinct performance gap, even the second efficient method NAAF \citep{naaf} requires 3.2$\times$ computations (0.5 vs 1.6 GFLOPs). Note that Drill-down \citep{drilldown} has the minimal model parameters yet requires comparatively higher GFLOPs. We argue the reason is that Drill-down \citep{drilldown} comprises two training stages: supervised pre-training and reinforcement learning (RL) optimization. Therefore, substantial computation resources are indispensable to facilitate the RL sampling policy. In summary, we argue that our model successfully strikes a balance between effectiveness and efficiency.

\subsection{Qualitative analysis}
In Fig. \ref{fig_9}, we show some qualitative examples with the top-1 ranked image region for each query (illustrated with the same color). From these results, we have several observations:

\begin{figure}[t]
        \centering
        \includegraphics[width=\columnwidth]{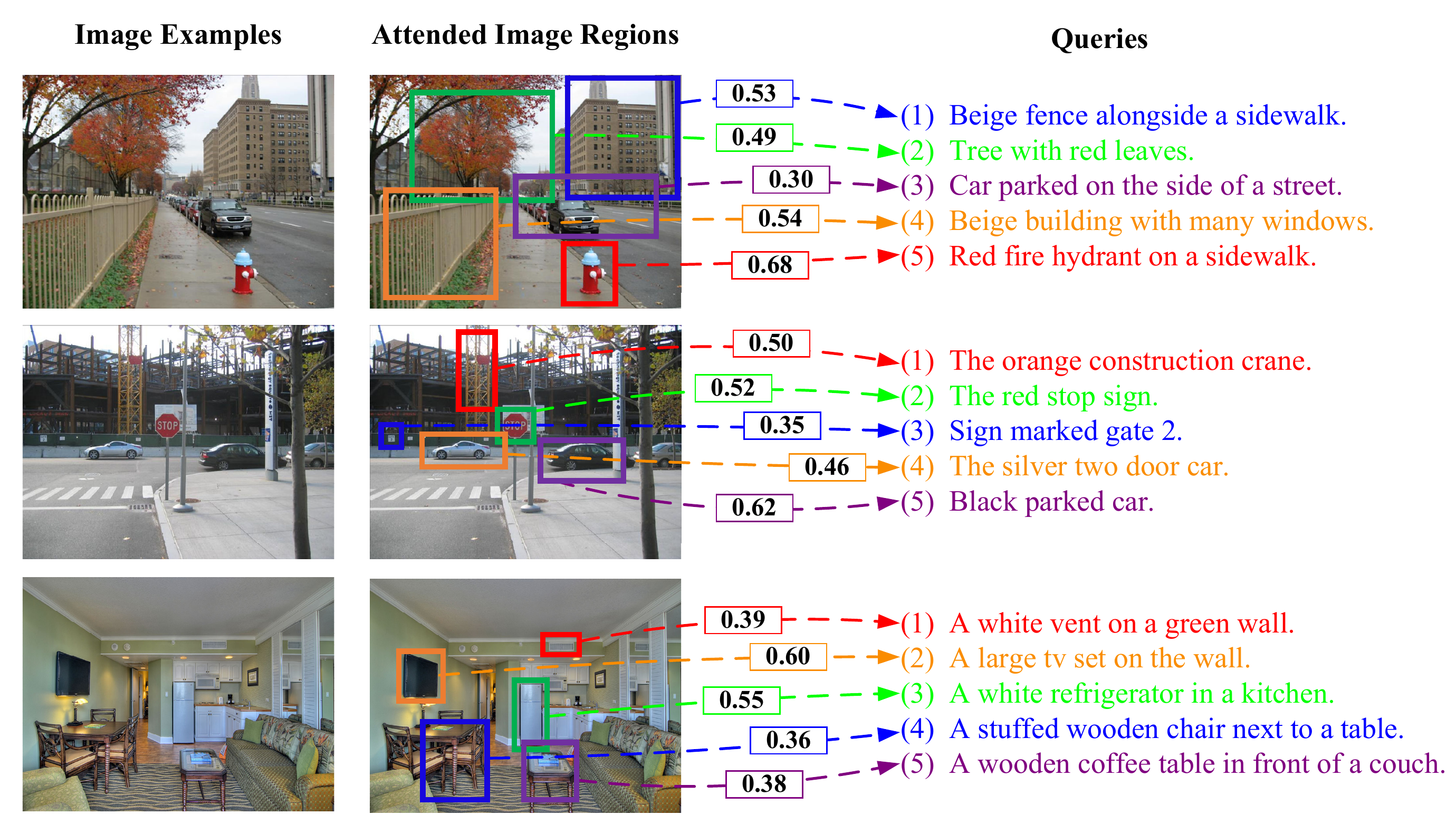}
        \caption{Qualitative retrieval examples of HMRN. Here the dash lines and arrows indicate the paired regions and queries. The first column shows image examples, and the second column displays the top-1 ranked salient region for each corresponding query. These attended regions are visualized with different colored bounding boxes, and the numbers on dash lines indicate the matching score between each salient region and its corresponding query.}
        \label{fig_9}
        \end{figure}

\begin{itemize}
        \item Our HMRN not only successfully retrieves the target image, but also focuses on the fine-grained alignment of salient image regions and region-specific queries. For each query, its paired region obtains the highest matching score among all potential image regions. For example, in the third image, the region indicated by the orange boundng box, which is in correspondence with the query ``A large tv set on the wall'', obtains the highest matching score. 
        \item The salient regions reveal the key semantic information of the image. Regions with higher similarity scores are usually more discriminative and contain unique clues. For instance, in the first image, the red bounding box paired with the query ``Red fire hydrant on a sidewalk'' is clearly different to other image regions, thus gaining a higher similarity. On the contrary, the query ``Car parked on the side of a street'' causes over-generalization and lacks certainty, resulting in a relatively lower score.
        \item Our model is capable to distinguish objects with similar captions, such as ``silver car'' and ``black parked car'' in the second image, ``wooden chair next to a table'' and ``wooden coffee table'' in the third image, which further verifies the necessity to explore the high-level semantic correlations among region-query pairs.
\end{itemize}

\section{Conclusion}
In this paper, we have proposed a novel Hierarchical Matching and Reasoning Network (HMRN) for Multi-Query Image Retrieval, which consists of two hierarchical modules: the Scalar-based Matching (SM) module and the Vector-based Reasoning (VR) module. Our designed hierarchical structure aligns the local and global semantic information, and meanwhile ensures the high-level correlation-enhanced correspondences. Specifically, the SM module obtains the scalar-based multi-level alignment similarities from the local level and the global level, and the VR module further explores the high-level reasoning similarity with strong semantic correlations, including the intra-correlation within each region-query pair and the inter-correlation among different region-query pairs. Finally, we aggregate above three-level similarities to form the ultimate similarity via the similarity aggregation strategy. We experimentally validate the advancement of our proposed HMRN on the Visual Genome dataset, which significantly exceeds the existing state-of-the-art methods. Meanwhile, the effect of each component in our model is testified in our ablation studies. 

In the future, considering that the hierarchical semantic representations are fused before adaptively estimating the effectiveness of each level, we aim to study an adaptive aggregation pattern to evaluate and flexibly adjust the aggregation weight of each level. We also plan to deploy our hierarchical matching and reasoning model on other complicated scenarios that require close attention to fine-grained details, e.g., scene text retrieval and cross-modal remote sensing image retrieval.


\printcredits

\section*{Declaration of competing interest}
The authors declare that they have no known competing financial interests or personal relationships that could have appeared to influence the work reported in this paper.

\section*{Acknowledgements}
This work was supported by the National Natural Science Foundation of China under Grants 62176178.

\bibliographystyle{unsrt}   

\bibliography{reference}



\end{document}